\documentclass[article]{elsarticle}
\usepackage{lineno,hyperref}

\usepackage{graphicx}
\usepackage{latexsym}
\usepackage{color}
\usepackage{caption}
\usepackage{subcaption}
\usepackage{bm}
\usepackage{bbm}
\usepackage{amsmath,amssymb, amsthm}
\usepackage{algorithm}
\usepackage{algpseudocode}
\usepackage{color,soul}
\usepackage{graphics}
\usepackage{amsfonts}
\usepackage{setspace}   
\usepackage{wasysym}
\usepackage{multirow}

\usepackage{adjustbox}
\usepackage{graphicx}
\usepackage{multirow}
\usepackage{nomencl}
\usepackage{float}
\usepackage[table]{xcolor}
\usepackage{standalone}
\usepackage{tikz}
\makenomenclature
\biboptions{sort&compress}
\newcommand*\diff{\mathop{}\!\mathrm{d}}
\newcommand*\Laplace{\mathop{}\!\mathbin\bigtriangleup}

\makeatletter
\newcommand{\thickhline}{%
	\noalign {\ifnum 0=`}\fi \hrule height 1pt
	\futurelet \reserved@a \@xhline
}
\newcolumntype{"}{@{\hskip\tabcolsep\vrule width 1pt\hskip\tabcolsep}}
\makeatother

\newcolumntype{M}[1]{>{\centering\arraybackslash}m{#1}}

\usepackage{geometry}
\makeatletter
\def\ps@pprintTitle{%
	\let\@oddhead\@empty
	\let\@evenhead\@empty
	\def\@oddfoot{\reset@font\hfil\thepage\hfil}
	\let\@evenfoot\@oddfoot
}
\makeatother

\begin{document}

\begin{frontmatter}

\title{A Deep Neural Network Surrogate for High-Dimensional~Random~Partial~Differential~Equations}

\author{Mohammad Amin Nabian}
\author{Hadi Meidani\corref{mycorrespondingauthor}}
\address{Department of Civil and Environmental Engineering, University of Illinois at Urbana-Champaign, Urbana, Illinois, USA.}
\cortext[mycorrespondingauthor]{Corresponding author}
\ead{meidani@illinois.edu}

\begin{abstract}
Developing efficient numerical algorithms for the solution of high dimensional random Partial Differential Equations (PDEs) has been a challenging task due to the well-known curse of dimensionality. We present a new solution framework for  these problems   based on a deep learning approach. Specifically, the random PDE is approximated by a feed-forward fully-connected deep residual network, with either strong or weak enforcement of initial and boundary constraints. The framework is mesh-free, and can handle irregular computational domains. Parameters of the approximating deep neural network are determined iteratively using variants of the Stochastic Gradient Descent (SGD) algorithm. The satisfactory accuracy of the proposed frameworks is numerically demonstrated on diffusion and heat conduction problems, in comparison with the converged Monte Carlo-based finite element results. 
\end{abstract}


\end{frontmatter}


\section{Introduction} \label{sec:introduction}
Partial differential equations (PDEs) are used to describe a variety of physical phenomena such as fluid dynamics, quantum mechanics, and elasticity. Reliable analysis of these phenomena often requires taking into account the inherent uncertainties in the system of interest (e.g. in their initial or boundary conditions, material properties, external forces) and quantifying the impact of these uncertainties on the quantities of interest (QoIs). Developing numerical methods for solving high-dimensional random PDEs, i.e. for systems with a large number of random parameters, has been a longstanding challenge. In this study, we propose a novel method based on deep neural networks for solving high-dimensional random PDEs. We consider the random parameters to be the only source of uncertainty in PDEs. 

Among the numerical methods proposed for random PDEs, Monte Carlo Sampling (MCS) is  one of the most commonly used method \cite{fishman2013monte}. In MCS, independent realizations from the random inputs are generated based on their prescribed probability distribution, leading to a number of deterministic PDE, each solved by a numerical method such as Finite Element (FE), Finite Difference (FD), and Finite Volume (FV). An ensemble of these deterministic solutions is then used in computing the response statistics. \emph{Perturbation methods} are another class of methods for random PDEs, where random outputs are expanded as a Taylor series around the means \cite{liu1986probabilistic}. These methods are suitable when the magnitude of uncertainty in the inputs and  outputs are expected to be small, and as such are limited in applicability. \emph{Operator-based methods} are another class which are based  on the manipulation of stochastic operators in PDEs, and include methods such as the weighted integral method \cite{deodatis1991weighted} and the Neumann expansion \cite{yamazaki1988neumann}. These methods are also limited to small uncertainties, and additionally, they are typically restricted to time-independent problems. \emph{Moment equations methods} aim to find the moments of the random solution to PDE directly, by deriving equations from the averages of the random PDEs. These methods suffer from the so-called closure problem \cite{zhang2001stochastic}. 

The \emph{Generalized Polynomial Chaos (gPC)} approach \cite{ghanem2003stochastic,xiu2010numerical,xiu2002wiener} is another widely used solutoin method for random PDEs. In gPC methods, the random variables (both the inputs and outputs) are represented as a linear combination of orthogonal polynomials of random parameters. The  gPC solution is generally obtained according to two different approaches of stochastic Galerkin \cite{ghanem2003stochastic} and stochastic collocation \cite{xiu2005high,doostan2011non,blatman2010adaptive}.  While the gPC provides a solid framework for solving random PDEs, they still suffer from a number of shortcomings. On the one hand, the stochastic Galerkin method is cumbersome to implement, and when the PDE takes highly complex and non-linear form, the explicit derivation of the Galerkin system is non-trivial. On the other hand, the stochastic collocation methods suffer from the curse of dimensionality and in high-dimensional systems the approximation errors can be significant, unless prohibitively large number of samples are obtained \cite{xiu2010numerical}.

Neural and deep neural networks have been proposed in previous studies for solving deterministic differential equations. These studies can generally be divided into two categories. In the first category, neural and deep neural networks are utilized in conjunction with other conventional methods (e.g. FE, FD, and FD) mainly to improve the computational efficiency of the existing methods. For instance, in \cite{lee1990neural,wang1990structured,gobovic1994analog,yentis1996vlsi}, the computation of difference equations that appear in the finite difference solution for a PDE was accelerated by neural networks. Specifically, the linear system of equations derived from the difference equations is mapped onto different neural network architectures, and the solution to the system of equations is obtained by minimizing the neural networks' energy functions. Also, the authors in \cite{meade1994solution,meade1994numerical} adopted linear B-splines to approximate the solution of  ordinary differential equations and used feed-forward neural networks  to determine the spline parameters. But, this method is not easily applicable for high dimensions \cite{lagaris1998artificial}. In \cite{tompson2016accelerating}, in an attempt to accelerate simulation of Navier-Stokes equations in an Eulerian framework, Convolutional Neural Networks (CNNs) were built to solve the sparse system of linear equations derived from the Poisson equation of pressure. In \cite{suzukineural}, an accelerated (in contrast to multi-scale simulation) approach is proposed based on regression analysis. Using a training dataset that is obtained from pre-analysis or experiments, local regression models are trained that represent the nonlinear relationship between the local nodal points. Each regression model then functions as a discretized equation for the PDE. Neural networks and polynomial functions are adopted as the regression model in this study.

In the second category, neural and deep neural networks are introduced as a substitute to the conventional methods. Generally in these methods, the solution to the deterministic PDE is represented in form of a neural or deep neural network. For instance in \cite{lagaris1998artificial}, neural networks are used to solve initial and boundary value problems. The solution is formed to consist  of two  parts, with the first part satisfying the initial or boundary conditions, and the second part being a neural network that is independent of the initial or boundary conditions. Parameters of the neural network are then calibrated by minimizing the squared residuals over specified collocation points. A somehow similar approach, called Deep Galerkin Method (DGM), was introduced in \cite{sirignano2017dgm}, with some key differences, namely (1) no trial function was used to strictly enforce the initial and boundary conditions, and these conditions were (weakly) enforced during the neural network training, (2) relatively deeper neural network architectures were used and training was performed with the state-of-the-art mini-batch optimization techniques, and (3) squared residuals were minimized iteratively over a set of randomly selected points per iteration, which makes the method mesh-less. Also, a hybrid variation of the approach proposed in \cite{lagaris1998artificial} is discussed in \cite{malek2006numerical}. In \cite{rudd2013solving},  the method of constrained backpropagation was used to train neural networks that can serve as the solution to nonlinear elliptic and parabolic PDEs. Also in  \cite{weinan2017deep,han2017overcoming}, a deep neural network solution approach was proposed for high-dimensional parabolic PDEs. To this end, PDEs are reformulated using backward stochastic differential equations and the solution to PDE is approximated by a neural network by making an analogy between the reinforcement learning and backward stochastic differential equations in which the gradient of the solution plays the role of policy function.

We present a framework for solving high-dimensional random PDEs  based on a deep-learning approach. This study is partially inspired by ideas in \cite{lagaris1998artificial,sirignano2017dgm}, where feed-forward neural and deep neural networks were used to solve high-dimensional deterministic differential equations. Two different approaches are introduced, which vary in the way the governing equation is satisfied, i.e. in the strong or variational form. In the first approach, similar in spirit to the Least-Square Finite Element Methods (LSFEM) \cite{bochev2006least}, we minimize the squared residual over the entire computational domain, with the random PDE solution represented in form of a feed-forward fully-connected deep residual network. In the second approach, we consider the variational form of the PDE of interest and approximate the solution to this weak form using feed-forward fully-connected deep residual networks. In both of the proposed algorithms, the deep neural network parameters are trained utilizing variants of the mini-batch gradient descent algorithm, and the solution is identified to satisfy either the strong or variational form of the PDE, and also the boundary and initial conditions. Random batches of spatial, temporal and stochastic points are sampled iteratively and for each sampled batch of points, the deep neural network parameters are updated by taking a descent step toward minimizing the loss function. Therefore, solving the random PDE is effectively reduced to solving an optimization problem. Similar to the methods in the second category discussed above, the proposed framework implements deep residual networks as a substitute to conventional methods, and solution to the random PDE is represented in form of a deep residual network.

The remainder of this paper is organized as follows. Deep neural and residual networks are introduced in Section \ref{sec:DNN}. A general algorithm for solving random PDEs using deep residual networks is introduced in Section \ref{framework}. Next, A modified algorithm based on variational formulation of PDEs is described in Section \ref{sec:variational}. Section \ref{sec:examples} represent three sample problems, including a diffusion process and two heat conduction problems, that are solved by utilizing the proposed method. Finally, the paper is concluded in Section \ref{sec:conclusion} with some discussion on the results, relative advantages and limitations of the proposed framework, and potential future works.

\section{Deep Neural Networks} \label{sec:DNN}

\subsection{Feed-Forward Fully-Connected Deep Neural Networks}
For notation brevity, let us first define the \textit{single hidden layer} neural network, since the generalization of the single hidden layer network to a network with multiple hidden layers, forming a \emph{deep} neural network, will be straightforward. Given the $d$-dimensional row vector $\bm{x} \in D^{d}$ as model input, the $k$-dimensional output of a standard single hidden layer neural network is in the form of
\begin{equation} \label{OHL-NN}
	\bm{y} = \sigma (\bm{x} \bm{W}_{1}+\bm{b}_{1}   ) \bm{W}_{2}+\bm{b}_{2},
\end{equation}
in which $\bm{W}_{1}$ and $\bm{W}_{2}$ are weight matrices of size $d\times q$ and $q\times k$, and $\bm{b}_{1}$ and $\bm{b}_{2}$ are \emph{bias} vectors of size $1\times q$ and $1\times k$, respectively. The function $\sigma( \cdot  )$ is an element-wise non-linear model, commonly known as the \textit{activation} function. In deep neural networks, the output of each activation function is transformed by a new weight matrix and a new bias, and is then fed to another activation function. Each new set of weight matrix and bias that is added to (\ref{OHL-NN}) constitutes a new \textit{hidden layer} in the neural network. Generally, the capability of neural networks to approximate complex nonlinear functions can be increased by adding more hidden layers or increase the dimensionality of the hidden layers. 

Popular choices of activation functions include Sigmoid, hyperbolic tangent (Tanh), and Rectified Linear Unit (ReLU). The ReLU activation function, one of the most widely used functions, has the form of $f( \theta  )=\max( 0,\theta  )$. However, second and higher-order derivatives of ReLUs is 0 (except at $\theta=0$). This limits its applicability in our present work which deals with differential equations consisting potentially of second or higher-order derivatives. Alternatively, Tanh or Sigmoid activations can be used for second or higher-order PDEs.

In a regression problem given a number of training data points, we may use a Euclidean loss function in order to calibrate the weight matrices and biases, as follows
\begin{equation} \label{MSE Loss}
	E_\textup{MSE}( \bm{X},\bm{Y})=\frac{1}{2M}\sum_{i=1}^{M}\left \| \bm{y}_i-\hat{\bm{y}}_{i} \right \|^{2},
\end{equation}
where $E_\textup{MSE}$ is the mean squared error, $\bm{X}=\left \{ \bm{x}_1,\bm{x}_2,...,\bm{x}_M \right \}$ is the set of $M$ given inputs, $\bm{Y}=\left \{ \bm{y}_1,\bm{y}_2,...,\bm{y}_M \right \}$ is the set of $M$ given outputs, and $\left \{ \hat{\bm{y}}_1,\hat{\bm{y}}_2,...,\hat{\bm{y}}_M \right \}$ is the set of neural network predicted outputs calculated at the same set of given inputs $\bm{X}$. 

The model parameters can be calibrated according to
\begin{equation} \label{minimize_loss}
	( \bm{W}_{1}^{*},\bm{W}_{2}^{*},\cdots,\bm{b}_{1}^{*},\bm{b}_{2}^{*},\cdots  )=\underset{{( \bm{W}_{1},\cdots,\bm{b}_{1}\cdots  )}}{\operatorname{argmin}} E_\textup{MSE}(\bm{X},\bm{Y}).
\end{equation}
Minimizing the loss function is usually performed using \emph{backpropagation} \cite{lecun2015deep}. In backpropagation, the gradients of an objective function with respect to the weights and biases of a deep neural network are calculated by starting off from the network output and propagating towards the input layer while calculating the gradients, layer by layer, using the chain rule. More details on feed-forward fully-connected deep neural networks can be found in \cite{lecun2015deep,goodfellow2016deep}. 

\subsection{Deep Residual Learning} 

It has been shown that depth in neural networks can in general make learning easier \cite{ba2014deep,simonyan2014very,srivastava2015training,srivastava2015highway,szegedy2015going}. The advantage of depth is that it can help learning features at various levels of abstraction. By increasing the number of layers, level of features can be enriched, giving deeper neural networks superiority in generalizability. However, deeper neural networks are more cumbersome to train, mainly due to the \emph{degradation} problem: it is observed numerically, that with an increase in  depth of neural networks, training and test accuracies get saturated and then degrade rapidly, possibly due to optimization issues \cite{he2016deep,he2015convolutional}.  

\begin{figure}
	\begin{center}
		\includegraphics[width=0.225\linewidth]{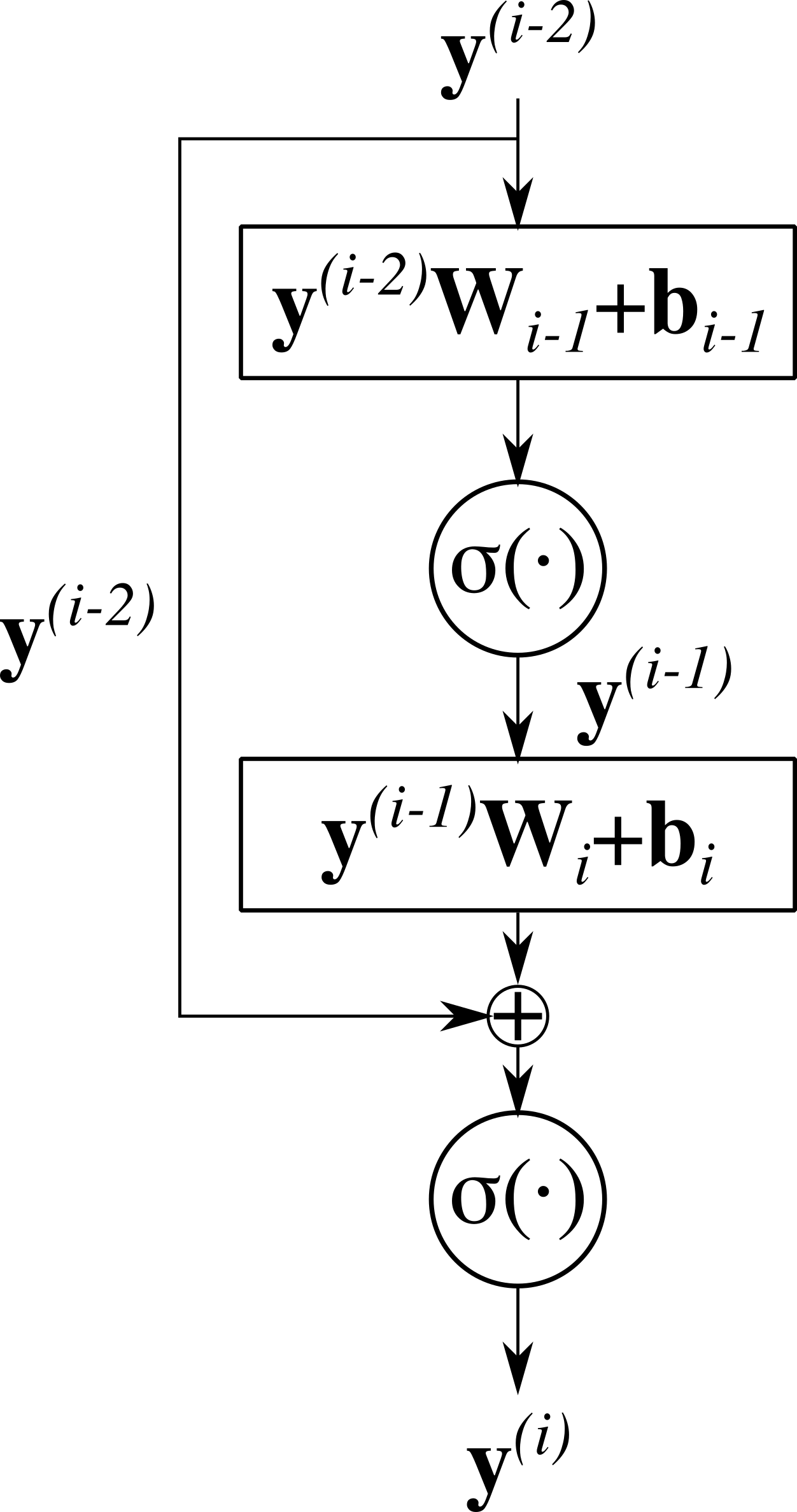}
		\caption{A sample deep residual network building block. $r$ is set to 2.} 
		\label{fig.resnet}
	\end{center}
\end{figure}

The degradation problem can be  addressed by the use of deep residual networks \cite{he2016deep}. Deep residual networks are  similar to feed-forward neural networks except that a number of shortcut connections are added to the architecture. In deep residual learning, the neural network is represented by a number of building blocks, each including one shortcut connection. A building block is defined as

\begin{equation} \label{building_block}
	\bm{y}^{(i)} = \sigma ( \mathcal{F} ( \bm{y}^{(i-r)}, \bm{W}_{i-r+1}, \bm{b}_{i-r+1}, \cdots, \bm{W}_{i}, \bm{b}_{i} ) +   \bm{y}^{(i-r)} ),
\end{equation}
where $r$ is the number of layers in a building block, $y^{(i)}$ is the output of layer $i$, and $\mathcal{F}( \cdot )$ is the residual mapping that takes  $y^{(i-r)}$ to the activation input in layer $i$. A sample building block is depicted in Fig.~\ref{fig.resnet} for $r=2$. Note that if we exclude the term $\bm{y}^{(i-r)}$ from the above equation, the equation essentially represents the mapping from the output of layer $i-r$ to the output of layer $i$ in a feed-forward deep neural network. For instance, consider a building block consisting of two layers right after the first layer of a deep neural network (i.e. $i=3, r=2$). Then the residual mapping can be expressed as

\begin{equation} \label{building_block}
	\mathcal{F} ( \bm{y}^{(i-r)}, \bm{W}_{i-r+1}, \bm{b}_{i-r+1}, \cdots, \bm{W}_{i}, \bm{b}_{i} ) = \sigma (\bm{y}^{(1)} \bm{W}_{2}+\bm{b}_{2}  ) \bm{W}_{3}+\bm{b}_{3}.
\end{equation}

The building block represented in Equation \ref{building_block} requires the output dimension of the residual mapping $\mathcal{F}( \cdot )$ be equal to the dimension of building block input $\bm{y}^{(i-r)}$. In cases where this condition does not hold, a linear projection may be applied to the input to building block to make the dimensions match \cite{he2016deep}, as follows

\begin{equation} \label{building_block_w/transformation}
	\bm{y}^{(i)} = \sigma ( \mathcal{F} ( \bm{y}^{(i-r)}, \bm{W}_{i-r+1}, \bm{b}_{i-r+1}, \cdots, \bm{W}_{i}, \bm{b}_{i} ) +  \bm{y}^{(i-r)} \bm{W}_s ),
\end{equation}
where $\bm{W}_s$ is a linear projector. 

Adding the shortcut connections to the feed-forward deep neural networks does not increase the number of tunable parameters or the computational complexity, and training can still be performed using the same optimizers, e.g., stochastic gradient descent. It is straightforward to implement the residual networks in common deep learning libraries, and there is no need to modify the solvers. It is shown in \cite{he2016deep} that very deep residual networks, compared to  the counterpart standard feed-forward deep neural networks,  are easier to optimize, do not suffer from the degradation problem as the depth increases, and enjoy the accuracy gains by increasing the depth of the network.

\section{Deep Learning for Random PDEs:  A Strong Form} \label{framework}

We seek to calculate the approximate  solution $u(t,\bm{x},\bm{p}; \bm{\theta} )$  for the following differential equation

\begin{equation}\label{eqn:pde}
	\begin{aligned}
		\mathcal{L}( t,\bm{x}, \bm{p} ; u (t,\bm{x},\bm{p}; \bm{\theta} ) ) =0, \; \; \; \; & t \in [ 0,T ], \bm{x}\in \mathcal{D},  \bm{p}\in \mathbb{R}^d, \\
		\mathcal{I}( \bm{x}, \bm{p} ; u (0,\bm{x},\bm{p}; \bm{\theta} ) )=0, \; \; \; \; & \bm{x}\in \mathcal{D}, \bm{p}\in \mathbb{R}^d,   \\
		\mathcal{B}( t,\bm{x}, \bm{p} ; u (t,\bm{x},\bm{p}; \bm{\theta} ) )=0, \; \; \; \;  & t \in [ 0,T ], \bm{x}\in \mathcal{\partial {D}}, \bm{p}\in \mathbb{R}^d, 
	\end{aligned}
\end{equation}
where $\theta$ include the parameters of the  function form of the solution,   $\mathcal{L}(\cdot)$ is a general differential operator that may consist of time derivatives, spatial derivatives, and linear and nonlinear terms, $\bm{x}$ is a position vector defined on a bounded continuous spatial domain $\mathcal{D} \subseteq \mathbb{R}^D , D \in \left \{ 1,2,3 \right \} $ with boundary $\mathcal{\partial {D}}$, $t \in \left[ 0,T \right]$, and $\bm{p}$ denotes an $\mathbb{R}^d$-valued random vector, with a joint distribution $\rho_{\bm{p}}$, that characterizes input uncertainties. Also, $\mathcal{I}(\cdot)$ and $\mathcal{B}(\cdot)$ denote, respectively, the initial and boundary conditions and may consist of differential, linear, or nonlinear operators. In order to calculate the solution, i.e. calculate the parameters  $\bm{\theta}$, let us consider the   following non-negative residuals, defined over the entire spatial, temporal and stochastic domains 
\begin{equation}\label{eqn:l2-redidual}
	\begin{aligned}
		r_\mathcal{L} (  \bm{\theta}  ) &=\int_{\left[ 0,T \right] \times \mathcal{D} \times \mathbb{R}^d  }( \mathcal{L} (  t,\bm{x}, \bm{p}; \bm{\theta} ) )^2 \rho_{\bm{p}}  \, \diff t \, \diff \bm{x} \, \diff \bm{p},   \\
		r_\mathcal{I} (  \bm{\theta}  ) &=\int_{\mathcal{D}\times \mathbb{R}^d  }( \mathcal{I} (  \bm{x}, \bm{p}; \bm{\theta} ) )^2 \rho_{\bm{p}}  \,\diff \bm{x} \, \diff \bm{p},   \\
		r_\mathcal{B} (  \bm{\theta}  )&=\int_{\left[ 0,T \right] \times \mathcal{\partial {D}}\times \mathbb{R}^d }( \mathcal{B} (  t,\bm{x}, \bm{p}; \bm{\theta} ) )^2 \rho_{\bm{p}}  \, \diff t \, \diff \bm{x} \, \diff \bm{p}.   
	\end{aligned}
\end{equation}
The optimal  parameters $\bm{\theta^*}$ can then be calculated according to 
\begin{equation} \label{eqn:theta-star}
	\begin{aligned}
		\bm{\theta^*}=\underset{{ \bm{\theta} }}{\operatorname{argmin}}\, r_\mathcal{L}( \bm{\theta}  ),   \\
		\text{s.t.} \quad r_\mathcal{I} (  \bm{\theta}  )=0, \, r_\mathcal{B} (  \bm{\theta}  )=0.
	\end{aligned}
\end{equation}
Therefore, the solution to the random differential equation defined in Equation \ref{eqn:pde} is reduced to an optimization problem, where initial and boundary conditions can   be viewed as constraints.  In this work, the constrained optimization~\ref{eqn:theta-star} is reformulated as an an unconstrained optimization with a modified loss function that also accommodate the constraints. To do so,  we adopt two different approaches, namely soft  and  hard assignment of constraints, which differ in  how strict the constraints are imposed \cite{marquez2017imposing}. In the soft assignment,  constraints are translated into additive penalty terms in the loss function (see e.g. \cite{sirignano2017dgm}). This approach is easy to implement but it is not clear how to  tune the relative importance of different terms in the loss function, and also there is no guarantee that the final solution will satisfy the constraints. In the hard assignment of constraints, the function form of the approximate solution  is formulated   in such a way that any solution with that function form is guaranteed to  satisfy the conditions (see e.g. \cite{lagaris1998artificial}). Methods with hard  assignment of constraints are more robust compared to their soft counterparts. However, the constraint-aware formulation of the function form of the solution is not straightforward for boundaries with irregularities or for mixed boundary conditions (e.g. mixed Neumann and Dirichlet boundary conditions). In what follows, we explain how the approximate solution in the form of a DNN can be calculated using these two assignment approaches. Let us denote the solution obtained by a feed-forward fully-connected deep residual network  by $u_{\text{DNN}}(t,\bm{x},\bm{p}; \bm{\theta} )$. The inputs to this deep residual network are $t$, $\bm{x}$, and realizations from the random vector $\bm p$.

For soft assignment of constraints, we use a generic DNN form for the solution. That is, we set $u_s(t,\bm{x},\bm{p}; \bm{\theta} ):=u_{\text{DNN}}(t,\bm{x},\bm{p}; \bm{\theta} )$, and solve the following unconstrained optimization problem
\begin{equation} \label{loss-soft}
	\bm{\theta^*}=\underset{{ \bm{\theta} }}{\operatorname{argmin}}\, \underbrace{r_\mathcal{L}(  \bm{\theta} ) +\lambda_1 r_\mathcal{I}  (  \bm{\theta}   )+\lambda_2 r_\mathcal{B}  (  \bm{\theta}   )}_{J_s(  \bm{\theta} ; u_s)},
\end{equation}
in which $\lambda_1$ and $\lambda_2$ are weight parameters, analogous to collocation finite element method in which weights are used to adjust the relative importance of each residual term \cite{bochev2006least}. 

In  hard assignment of constraints, the uncertainty-aware function form for the approximate solution can take the following general form \cite{lagaris1998artificial}.

\begin{equation} \label{trial-function}
	u_h (t,\bm{x},\bm{p}; \bm{\theta} )=C( t, \bm{x})+G( t,\bm{x},u_{\text{DNN}}(t,\bm{x},\bm{p}; \bm{\theta} )),
\end{equation}
where $C ( t, \bm{x} )$ is a function that satisfies the initial and boundary conditions and has no tunable parameters, and, by construction, $G( t,\bm{x},u_{\text{DNN}} (t,\bm{x},\bm{p}; \bm{\theta}  ) )$ is derived such that it has no contribution to the initial and boundary conditions. A systematic way to construct the functions $C(  \cdot )$ and $G(  \cdot)$ is presented in \cite{lagaris1998artificial}.  Our goal is then to estimate the DNN parameters $\bm \theta$ according to 
\begin{equation} \label{loss-hard}
	\bm{\theta^*}=\underset{{ \bm{\theta} }}{\operatorname{argmin}}\, \underbrace{r_\mathcal{L} (  \bm{\theta}  )}_{J_h(  \bm{\theta};u_h )}.
\end{equation}

To solve the two unconstrained optimization problems ~\ref{loss-soft} and~\ref{loss-hard}, we make use of stochastic gradient descent (SGD) optimization algorithms \cite{ruder2016overview}, which are a variation of gradient-descent algorithms. In each iteration of an SGD algorithm,  the gradient of loss function is approximated using only one point in the input space, based on which the neural network parameters are updated. This iterative update is shown to result in an unbiased estimation of the gradient, with bounded variance \cite{bottou2010large}.

Specifically, in soft assignment of constraints, on the $i^{\textit{th}}$ iteration, the DNN parameters are updated according to
\begin{equation} \label{ldescent step-soft}
	\bm{\theta}^{(i+1)} = \bm{\theta}^{(i)} - \eta^{(i)} \nabla_{\bm{\theta}}\tilde{J}_s^{(i)}(\bm{\theta}),
\end{equation}
where $\eta^{(i)}$ is  the step size in the $i^{\textit{th}}$ iteration, and $\tilde{J}_s^{(i)}(\bm{\theta})$ is the  approximate loss function, obtained by numerically evaluating integrals in Equations~\ref{eqn:l2-redidual} using a single   sample point. That is, 
\begin{multline} \label{loss-approximate-soft}
	\tilde{J}_s^{(i)}(\bm{\theta})= \left[\mathcal{L}( t^{(i)},\bm{x}^{(i)},\bm{p}^{(i)}; u_s(t^{(i)},\bm{x}^{(i)},\bm{p}^{(i)} ;\bm{\theta} ) ) \right]^2 + \\
	\lambda_1 \left[\mathcal{I}( \bm{x}^{(i)},\bm{p}^{(i)}; u_s(0,\bm{x}^{(i)},\bm{p}^{(i)} ;\bm{\theta} ) ) \right]^2 
	+\lambda_2 \left[\mathcal{B}( t^{(i)},\bm{\underbar{x}}^{(i)},\bm{p}^{(i)}; u_s(t^{(i)},\underbar{$\bm{x}$}^{(i)},\bm{p}^{(i)} ;\bm{\theta} ) ) \right]^2.
\end{multline}
where $t^{(i)},\bm{x}^{(i)}$ and $\underbar{$\bm{x}$}^{(i)}$ are  uniformly drawn in $\left[ 0,T \right], \mathcal{D}$ and $\mathcal{\partial{D}}$, and $\bm{p}^{(i)}$ is drawn in  $\mathbb{R}^d $ according to distribution $\rho_{\bm{p}}$. The gradient of loss function with respect to model parameters $\nabla_{\bm{\theta}}\tilde{J}_s^{(i)}(\bm{\theta})$ is calculated using backpropagation \cite{lecun2015deep}, which is a especial case of the more general technique called reverse-mode automatic differentiation \cite{baydin2015automatic}. In backpropagation, gradient calculation proceeds backwards through the deep neural network. The gradient of the last layer is calculated first and the gradient of the first layer is calculated last. Partial gradient computations for one layer are reused in the gradient computations for the foregoing layers. This backward flow of information facilitates efficient computation of the gradient at each layer of the deep neural network \cite{lecun2015deep}. The term $\mathcal{L}( t^{(i)},\bm{x}^{(i)},\bm{p}; u_s(t^{(i)},\bm{x}^{(i)},\bm{p}^{(i)} ;\bm{\theta})) $ also involves gradients of the solution $u_s$ with respect to $t$ and $\bm{x}$, which are calculated using reverse-mode automatic differentiation.

Similarly, in hard assignment of constraints, the DNN parameters are updated according to 
\begin{equation} \label{ldescent step-hard}
	\bm{\theta}^{(i+1)} = \bm{\theta}^{(i)} - \eta^{(i)} \nabla_{\bm{\theta}}\tilde{J}_h^{(i)}(\bm{\theta}),
\end{equation}
where
\begin{equation} \label{loss-approximate-hard}
	\tilde{J}_h^{(i)}(\bm{\theta}  )=\left[\mathcal{L}( t^{(i)},\bm{x}^{(i)},\bm{p}^{(i)}; u_h(t^{(i)},\bm{x}^{(i)},\bm{p}^{(i)} ;\bm{\theta}  ) ) \right]^2.
\end{equation}

It is common in the practice that in each iteration the gradient of the loss function is calculated at and averaged over $n$ different sample input points instead of being evaluated at only one point. Such approaches are called mini-batch gradient descent algorithms \cite{ruder2016overview}, and compared to stochastic gradient descent algorithms, are more robust and more   efficient. 

Algorithm \ref{Algorithm} summarizes the proposed step-by-step approach. 
The algorithm can be stopped based on a pre-specified  maximum number of iterations (as shown in Algorithm~\ref{Algorithm}, or using an on-the-fly stoppage criteria based on variations in  the loss function values across a few iterations.

\begin{algorithm}[H]
	\caption{Deep learning algorithm for random differential equations}\label{Algorithm}
	\begin{algorithmic}[1]
		\State Set the DNN architecture (number of layers, dimensionality of each layer, and activation function; and for residual networks, also the structure of shortcut connections).
		\State Initialize DNN parameters $\bm{\theta}^{(0)}$.
		\State Select a method $m$, with $m \in \{h,s\}$ (hard or soft assignment of constraints).
		\State Form the target function $u_h (t,\bm{x},\bm{p}; \bm{\theta} )$ according to Equation \ref{trial-function}.
		\State Form the loss function $\tilde{J}_h( \bm{\theta} )$ or $\tilde{J}_s( \bm{\theta} )$ according to Equations \ref{loss-approximate-hard} or \ref{loss-approximate-soft}.
		\State For the mini-batch gradient descent algorithm, specify optimizer hyper-parameters and batch size $n$.
		\State Specify  maximum number of iterations $i_{\text{max}}$; set $i=0$.
		\While {${i<i_{\text{max}}}$ }
		\If {m=h}
		\State Generate $n$ random input points $\{ t_j^{(i)},\bm{x}_j^{(i)},\bm{p}_j^{(i)} \}_{j=1}^{n}$, sampled uniformly from $[0,T] \times \mathcal{D}$, and from $\mathbb{R}^d $ according to $\rho_{\theta}$.
		\State Calculate the loss function $\tilde{J}_h^{(i)}( \bm{\theta} )$.
		\State Take a descent step  $\bm{\theta}^{(i+1)} = \bm{\theta}^{(i)} - \eta^{(i)} \nabla_{\bm{\theta}}\tilde{J}_h^{(i)}$.
		\Else
		\State Generate $n$ random input points $\{ t_j^{(i)},\bm{x}_j^{(i)},\bm{p}_j^{(i)} \}_{j=1}^{n}$, sampled uniformly from $[0,T] \times \mathcal{D}$, and from $\mathbb{R}^d $ according to $\rho_{\theta}$. Also, generate $n$ random inputs $\{\bm{\bar{x}}_j^{(i)}\}_{j=1}^{n}$ uniformly  from $\mathcal{\partial{D}}$.
		\State Calculate the loss function $\tilde{J}_s^{(i)}( \bm{\theta} )$.
		\State Take a descent step  $\bm{\theta}^{(i+1)} = \bm{\theta}^{(i)} - \eta^{(i)} \nabla_{\bm{\theta}}\tilde{J}_s^{(i)}$.
		\EndIf
		\State $i = i+1$
		\EndWhile\
		
	\end{algorithmic}
\end{algorithm}

\section{Deep Learning for Random PDEs:  A Variational Form and a Lower Order Loss Function} \label{sec:variational}
In this section, we  present a modified  algorithm which can be computationally more robust and more efficient. In particular, we address the numerical difficulties that arise when the operator $\mathcal{L}$ in Equation \ref{eqn:pde}  includes second or higher-order  derivatives, which are more expensive to compute when the response is represented by a deep neural networks \cite{sirignano2017dgm}, and could potentially lead to ill-conditioned Hessians \cite{saarinen1993ill}. To address this issue,  we propose an approach which  derives a first-order variational form of the operator  $\mathcal{L}$.

It should be noted that high order differentiation are typically with respect to spatial coordinates, and therefore in what follows, we show the derivation with regards to the integral over the spatial domain, and omit the dependences on time and random inputs for the sake of notation brevity. A typical approach to decrease the differentiation order is the use of variational form, where  instead of solving the original PDE in Equations \ref{eqn:pde} we solve the following variational form
\begin{equation} \label{variational_v}
	\int_{\mathcal{D}} \mathcal{L}(  \bm{x} ; u( \bm{x}  ) ) \ v( \bm{x}  ) \  \diff \bm{x}  =0, \quad \forall v(\bm{x}) \in \mathcal{V},
\end{equation}
where $v(\bm{x} )$ are referred to as `test functions', and  $ \mathcal{V}$ is the space of solutions that conform with the initial and boundary conditions (see e.g. \cite{boffi2013variational}). The order of spatial derivatives in $\mathcal{L}$  is then reduced by using integration by parts successively in Equation \ref{variational_v}, which leads to
\begin{equation} \label{variational}
	\int_{\mathcal{D}}\hat{\mathcal{L}}( \bm{x}; u(\bm{x} ),v(\bm{x} ) ) \  \diff \bm{x}  =c(\bm{x}; u(\bm{x} )), \quad \forall v \in \mathcal{V},
\end{equation}
where $\hat{\mathcal{L}}$ is a general differential operator that will consist of spatial derivatives that are of first order, and potentially other linear and nonlinear terms, and time derivatives. Also, the function $c(\cdot)$ is pertinent to the surface integration term, derived from the integration by parts,  usually in the form of a constant  for simple Dirichlet or Neumann boundary conditions. Therefore, in these cases we will have the following generic form
\begin{equation} \label{variational:noc}
	\int_{\left[ 0,T \right]   \times \mathbb{R}^d  } \int_{\mathcal{D}} \hat{\mathcal{L}}( t,\bm{x},\bm{p}; u(t,\bm{x},\bm{p} ),v(t,\bm{x},\bm{p} ) ) \ \diff \bm{x} \ \rho_{\bm{p}}  \ \diff t  \, \diff \bm{p} =0, \quad \forall v \in \mathcal{V}.
\end{equation}
In this study we are representing the solution $u$ by a DNN whose parameters are identified in a minimization problem, we need to find the new `low order' loss function $r_l(\theta)$ whose extremum is also the solution to Equation \ref{variational}. In other words,  let us consider the following (unknown) loss function
\begin{equation} \label{loworderloss}
	r_l(u) = \int_{\left[ 0,T \right] \times \mathcal{D} \times \mathbb{R}^d  } \mathcal{F}( t,\bm{x},\bm{p}; u(t,\bm{x},\bm{p}; \bm \theta ) ) \ \rho_{\bm{p}} \ \diff t \diff \bm{x} \diff \bm{p}=0. 
\end{equation}
Recalling calculus of variation, our objective is to identify the integrand $\mathcal{F}$ such that its extremum $u^*$ that satisfies  the following weak form (based on the first variation)
\begin{equation} \label{weakform}
	\frac{\diff}{\diff \epsilon} r_l( u^*+\epsilon v)\Big\rvert_{\epsilon = 0} =  \lim_{\epsilon\to 0}  \frac{1}{\epsilon}  \int_{\left[ 0,T \right] \times \mathcal{D} \times \mathbb{R}^d  }    [{\mathcal{F}( u^*+\epsilon v )-\mathcal{F}( u^* )}] \ \rho_{\bm{p}} \ \diff t \diff \bm{x} \diff \bm{p}=0, \quad \forall v \in \mathcal{V},
\end{equation}
also satisfies the variational form~\ref{variational}. In other words, after deriving the variational form~\ref{variational}, we set that equivalent to the weak form of~\ref{weakform}, and derive the integrand $\mathcal{F}$, accordingly. That is, we solve the following equation for $\mathcal{F}$
\begin{equation} \label{weakform}
	\lim_{\epsilon\to 0}  \frac{1}{\epsilon}     [{\mathcal{F}( u^*+\epsilon v )-\mathcal{F}( u^* )}]  = \hat{\mathcal{L}}( t,\bm{x},\bm{p}; u(t,\bm{x},\bm{p} ),v(t,\bm{x},\bm{p} ) ), \quad \forall v \in \mathcal{V}.
\end{equation}
It should be noted that in this process, the  derivatives in $\mathcal{F}$ will remain first order, as the process  involves only first variation and  only linear terms are kept. Appendix \ref{app:poisson} includes a sample derivation for a Poisson equation together with general tips on  how to derive the lower order loss functions. Considering a DNN form for solution, $u_\text{DNN}(t,\bm{x},\bm{p}; \bm{\theta} )$, the lower order loss function will be effectively parameterized by the  DNN parameters $\bm{\theta^*}$ leading to the following minimization problem to be solved by the SGD algorithm (using either soft or hard assignment of constraints),
\begin{eqnarray}\label{eqn:theta-star_variational}
	&\bm{\theta^*}=\underset{{ \bm{\theta} }}{\operatorname{argmin}}\, r_l( \bm{\theta}  ),  \nonumber \\
	& \text{s.t.} \, \, r_\mathcal{I} (  \bm{\theta}  )=0, \, r_\mathcal{B} (  \bm{\theta}  )=0.
\end{eqnarray}

\section{Numerical Examples} \label{sec:examples}

In this section, we numerically study the performance of the proposed deep learning method in solving sample high-dimensional random PDEs. In the first example, the proposed method is applied to solve random transient diffusion processes. Next, a steady heat conduction example is considered. Finally, the performance of the proposed method with soft assignment of constraints is examined by solving the heat conduction equation on a somewhat irregular domain. Training is performed in parallel across 4 NVIDIA P100 GPUs. The number of layers for the surrogates trained in examples 1, 2, and 3 are, respectively 24, 20, and 20. Dimensionality of each hidden layer is set to 256. Consequent odd hidden layers are connected to each other via a shortcut connection, as shown in figure \ref{fig.resnet}, with $r=2$. Tanh nonlinearities are adopted for each hidden layer. The Adam optimization algorithm \cite{kingma2014adam} is used to find the optimal surrogate parameters, mini-batch size is set to 32, and the learning rates, $\beta_1$, $\beta_2$, and $\epsilon$ for the Adam optimizer are set to $10^{-5}$, 0.9, 0.999, and $10^{-15}$, respectively. Number of epochs is set to $10^8$. All the examples are also solved by the Monte Carlo Finite Element (MC-FE) method (using MATLAB Partial Differential Equation Toolbox), and results are compared with each other. The MC-FE results are obtained by repetitively drawing samples from the random field, at which the finite element solver is run, until the convergence of the QoI is achieved.

\subsection{Transient diffusion problem with random coefficient} \label{sec:example1}

In the first example, we consider the transient diffusion problem in one-dimensional spatial domain with random diffusion property. Specifically, the governing PDE is given by

\begin{align}\label{eqn:diffusion}
	& \frac{\partial{u( t,x, \bm{p} )}}{\partial{t}} - \nabla \cdot ( a( x,\bm{p} ) \nabla u( t,x, \bm{p} ) )=c,\; \; \;  t\in\left [ 0,1 \right ],x\in\left [ 0,1 \right ],\bm{p} \in \left [ -1,1 \right ]^{d} \nonumber \\
	& u( t,0,\bm{p} )=0, \, u( t,1,\bm{p} )=0, \, u( 0,x,\bm{p} )= 10(x-x^2),
\end{align}
where $c=3$,  $a( x,\bm{p} )$ is the diffusion coefficient characterized  using the  $d$-dimensional  random vector $\bm{p}=\left \{ p_1,\cdots, p_{d} \right \}$. We consider two  different analytical forms for $a( x,\bm{p} )$, as will be discussed later. In both cases, following a hard assignment approach,  we set the deep neural network surrogate to take the following form

\begin{equation} \label{trial-function-example1}
	u(t,x,\bm{p}; \bm{\theta} )=10(x-x^2)+t(x-x^2)u_\text{DNN}(t,x,\bm{p}; \bm{\theta} ).
\end{equation}

\subsubsection{A smooth random field for the diffusion coefficient} 

\begin{figure}
	\begin{center}
		\begin{subfigure}[t]{0.44 \linewidth}
			\includegraphics[width=0.99\linewidth]{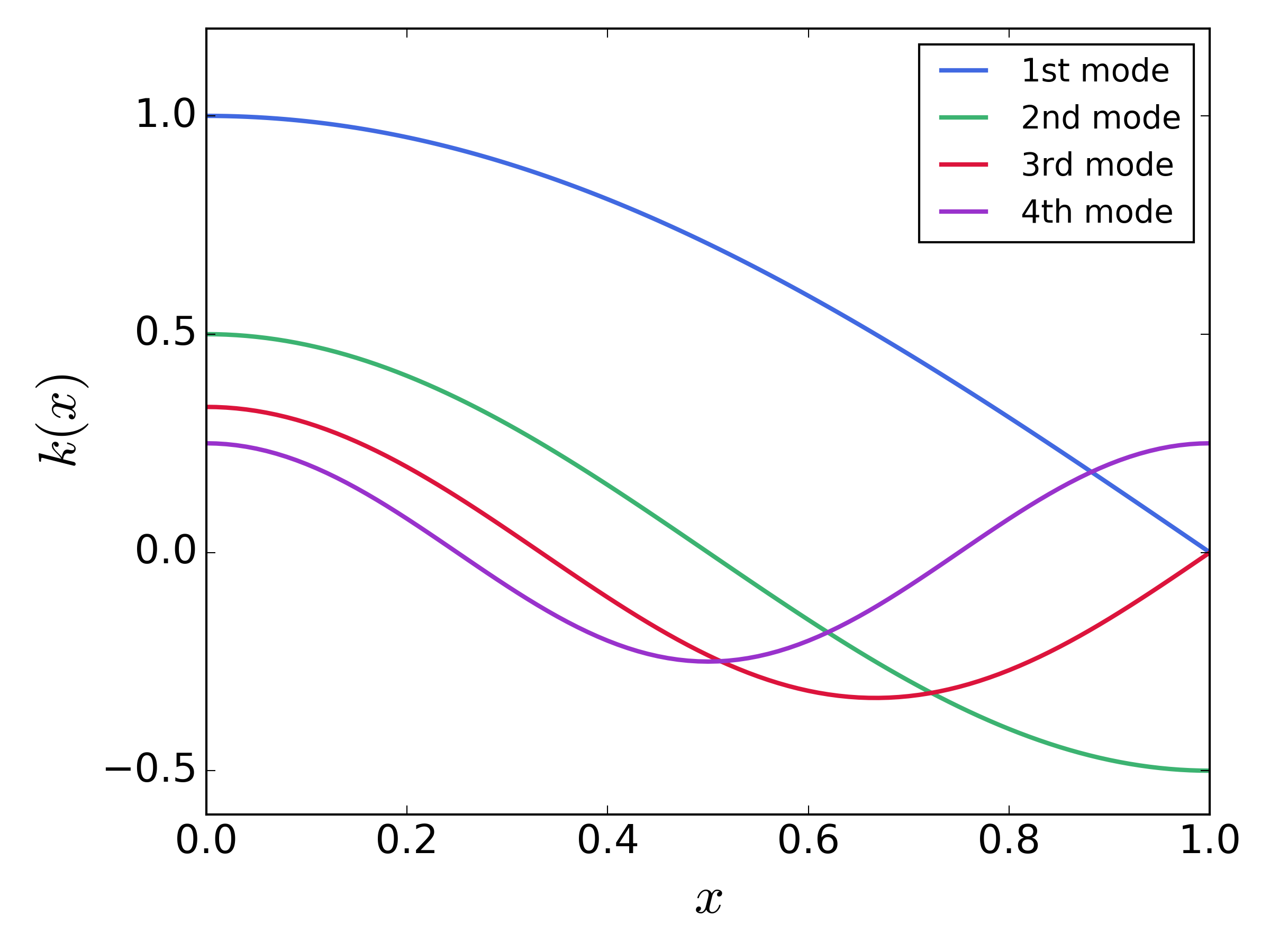}
			\caption{} 
			\label{fig.diffusion1_modes}	
		\end{subfigure}
		\quad
		\begin{subfigure}[t]{0.44 \linewidth}
			\includegraphics[width=0.99\linewidth]{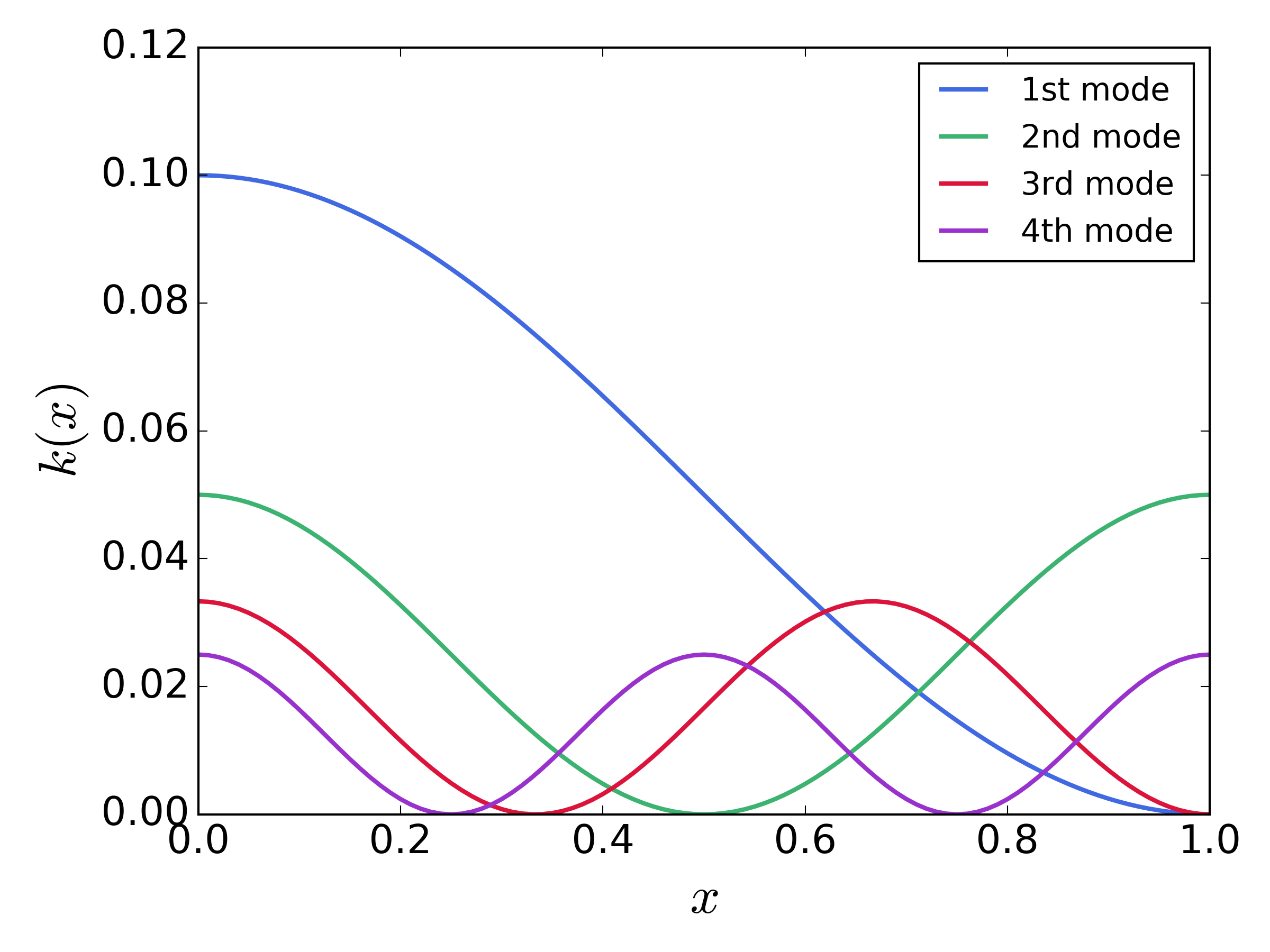}
			\caption{} 
			\label{fig.diffusion2_modes}
		\end{subfigure}
		\captionsetup{}
	\end{center}
	\caption{The first four spatial modes of the random field $a( x,\bm{p} )$, associated with (a) the smooth field represented by Equation \ref{eqn:diffusion_coeff_1}, and (b) the non-smooth field represented by Equation \ref{eqn:diffusion_coeff_1}. } 
	\label{fig.diffusion_pdf}
\end{figure}

\begin{figure}
	\begin{center}
		\includegraphics[width=0.88\linewidth]{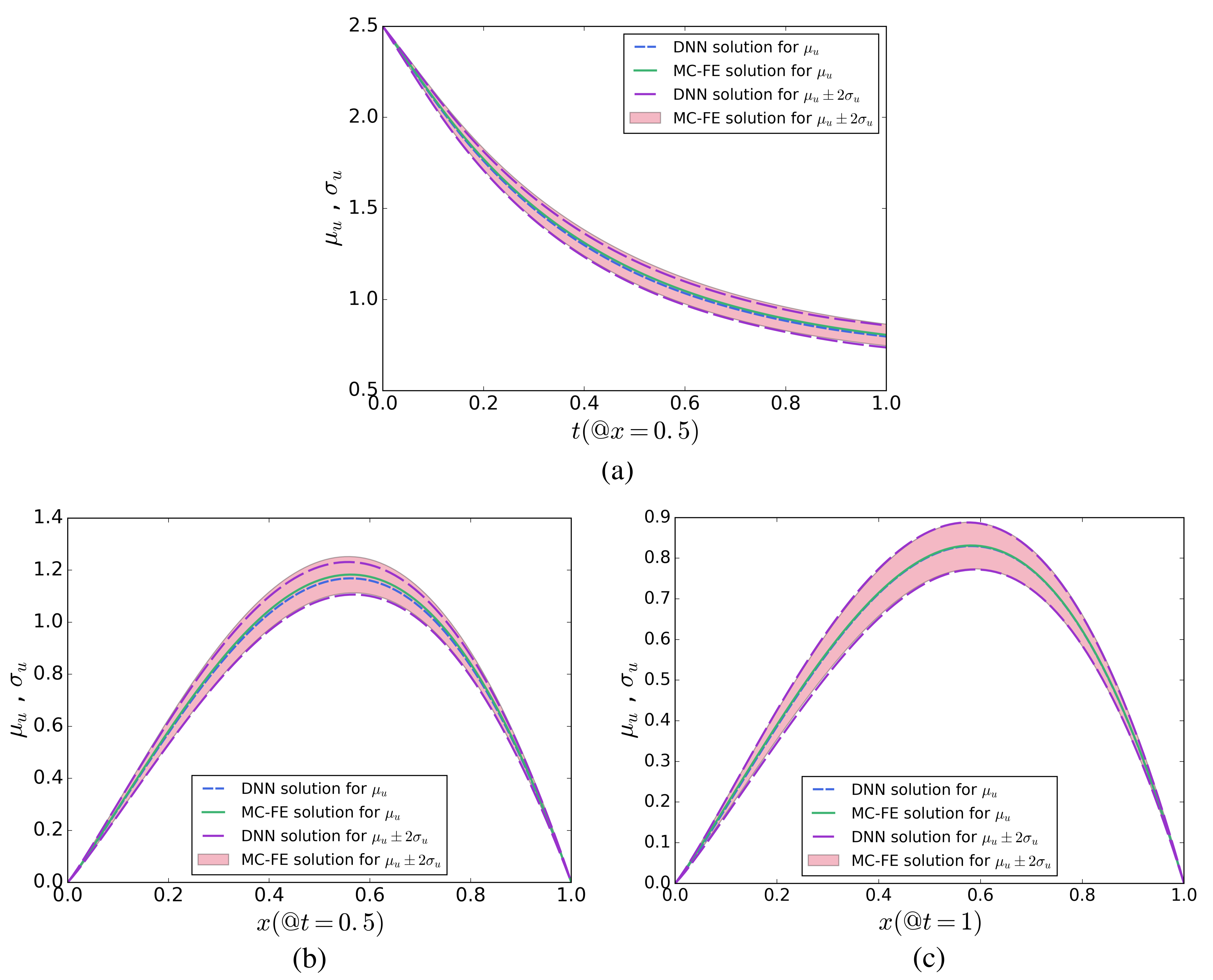}
		\caption{A comparison between the DNN and MC-FE results for the mean and standard deviation of $u$ at (a) $x=0.5$, (b) $t=0.5$, and (c) $t=1$ . The diffusion coefficient is represented by Equation \ref{eqn:diffusion_coeff_1}.}
		\label{fig.diffusion1_stats}
	\end{center}
\end{figure}

In this part, we  consider the diffusion coefficient to be represented by

\begin{equation}\label{eqn:diffusion_coeff_1}
	a( x,\bm{p} )=0.26+\sum_{j=1}^{d}\frac{0.05}{j}\cos( \frac{\pi}{2} jx )p_j,
\end{equation}
where   the random variables $\left \{ p_j \right \}_{j=1}^{d}$ are independent identically distributed according to uniform distribution on $\left [ 0,1 \right ]$, and  $d=100$. The first four modes of this random field are depicted in figure \ref{fig.diffusion1_modes}. Figure \ref{fig.diffusion1_stats} shows the satisfactory agreement between the means and standard deviations of $u$ obtained from the DNN framework and the MC-FE method.  For the FE solver, the mesh edge length is set to 0.005, which resulted in a total of 201 nodes, the size of time steps is  0.01 seconds, and a  total of $10^4$ MC samples, observed to be sufficient for the calculation of mean and standard deviation of the response, are used.

\subsubsection{A non-smooth random field for the diffusion coefficient} 

\begin{figure}
	\begin{center}
		\includegraphics[width=0.88\linewidth]{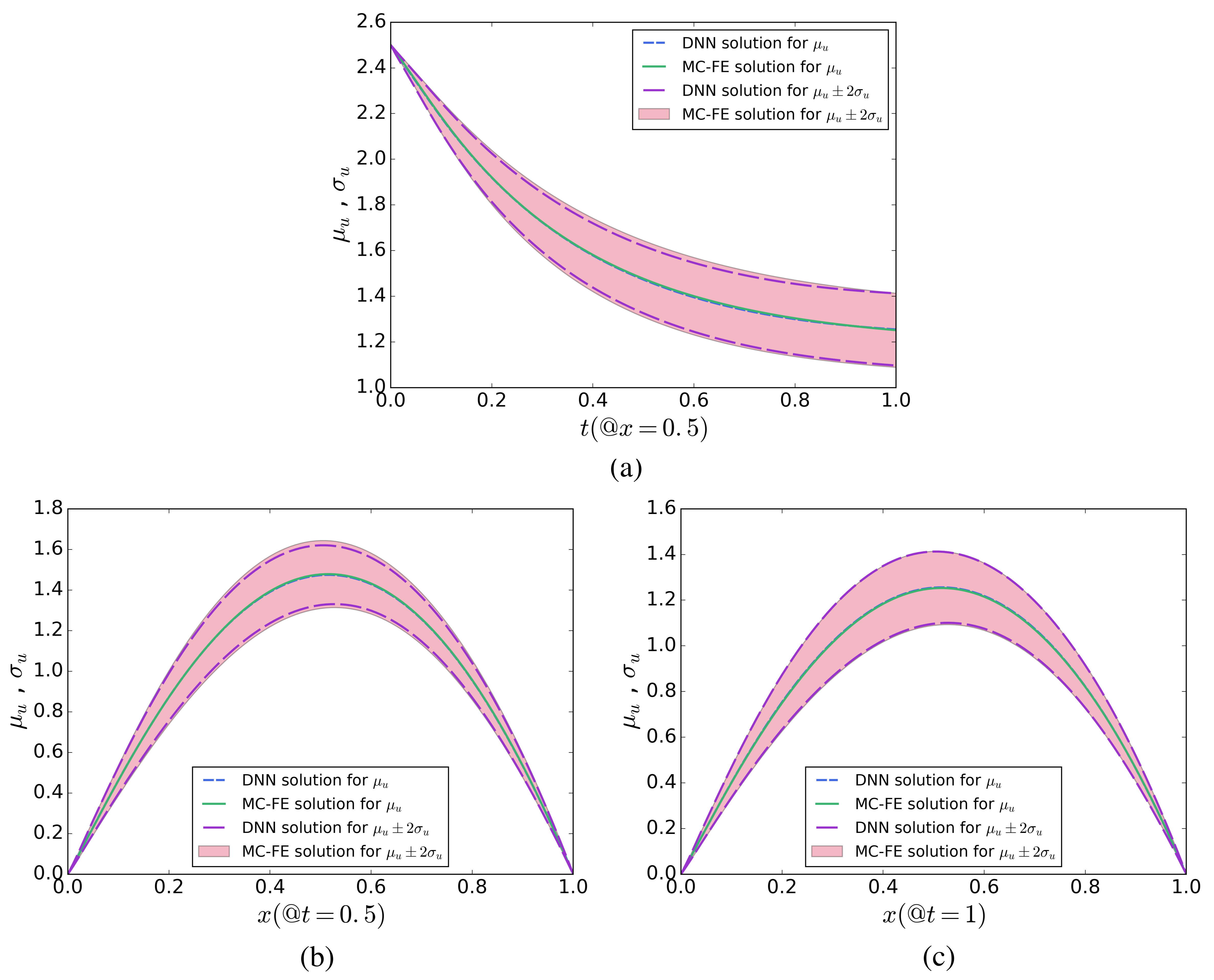}
		\caption{A comparison between the DNN and MC-FE results for the mean and standard deviation of $u$ at (a) $x=0.5$, (b) $t=0.5$, and (c) $t=1$. The diffusion coefficient is represented by Equation \ref{eqn:diffusion_coeff_2}.} 
		\label{fig.diffusion2_stats}
	\end{center}
\end{figure}

Here, we assume the diffusion coefficient to take the following form

\begin{equation}\label{eqn:diffusion_coeff_2}
	a( x,\bm{p} )=0.2+\sum_{j=1}^{d}\frac{0.1}{j}\cos^2( \frac{\pi}{2} jx )p_j,
\end{equation}
where $d=50$. The first four modes of this random field are depicted in figure \ref{fig.diffusion2_modes}, showing that this random field  is less smooth compared to the  one in the previous case. As can be seen in Figure \ref{fig.diffusion2_stats},  a good agreement exists between the DNN and MC-FE results, in terms of  the means and standard deviations of $u$. A FE model similar to the one detailed in the previous example was created.

Figure \ref{fig.diffusion_pdf} shows a comparison between the deep neural network and finite element results for the probability density function of $u(t=1,x=0.5)$. The plot on the left and right, respectively, correspond to the solutions computed in parts 1 and 2. It is evident that there is a good agreement between the results.

\begin{figure}
	\begin{center}
		\begin{subfigure}[t]{0.44 \linewidth}
			\includegraphics[width=.99\linewidth]{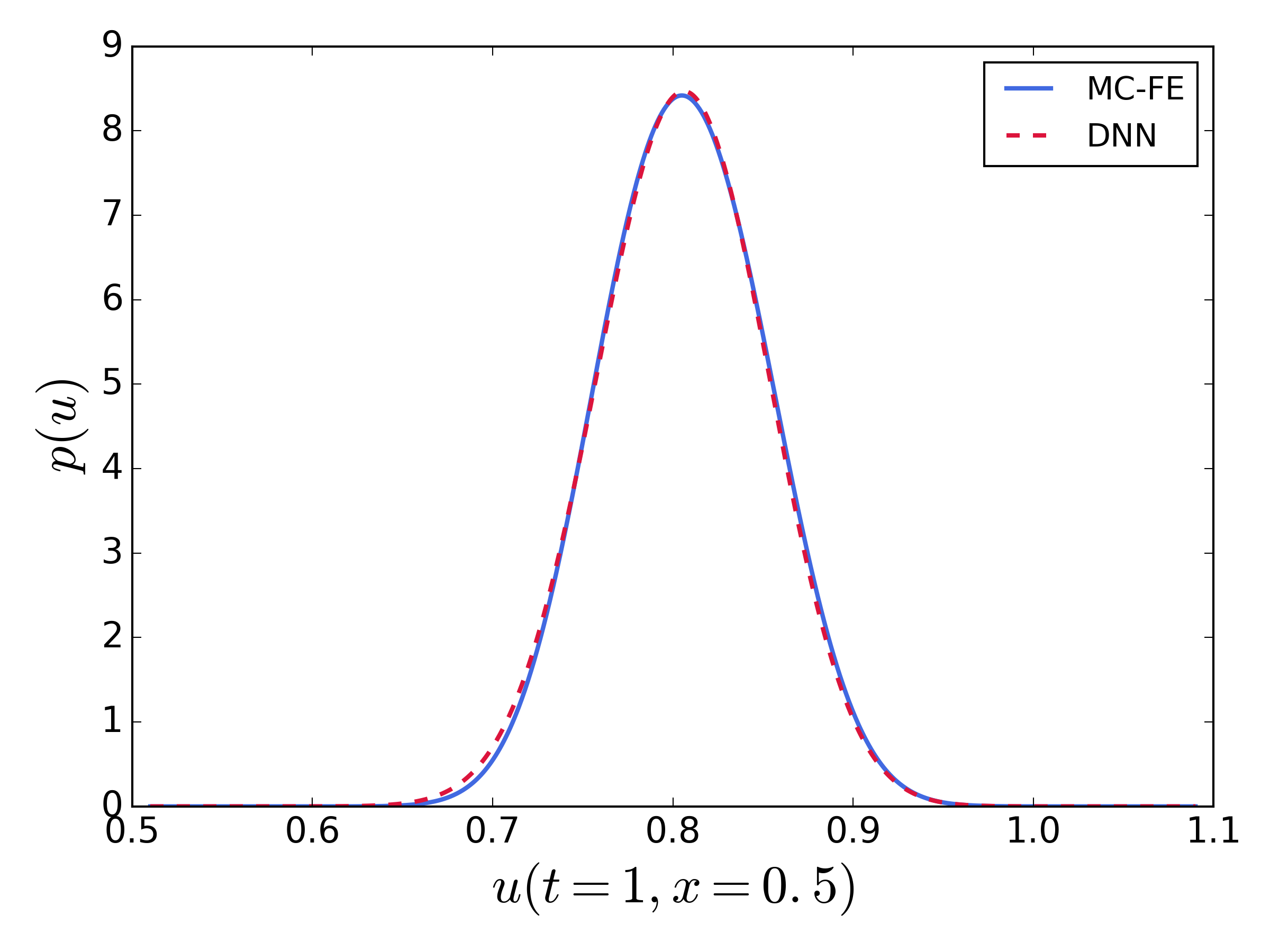}
			\caption{}
			\label{fig:diffusion1_pdf}	
			\centering	
		\end{subfigure}
		\quad
		\begin{subfigure}[t]{0.44 \linewidth}
			\includegraphics[width=.99\linewidth]{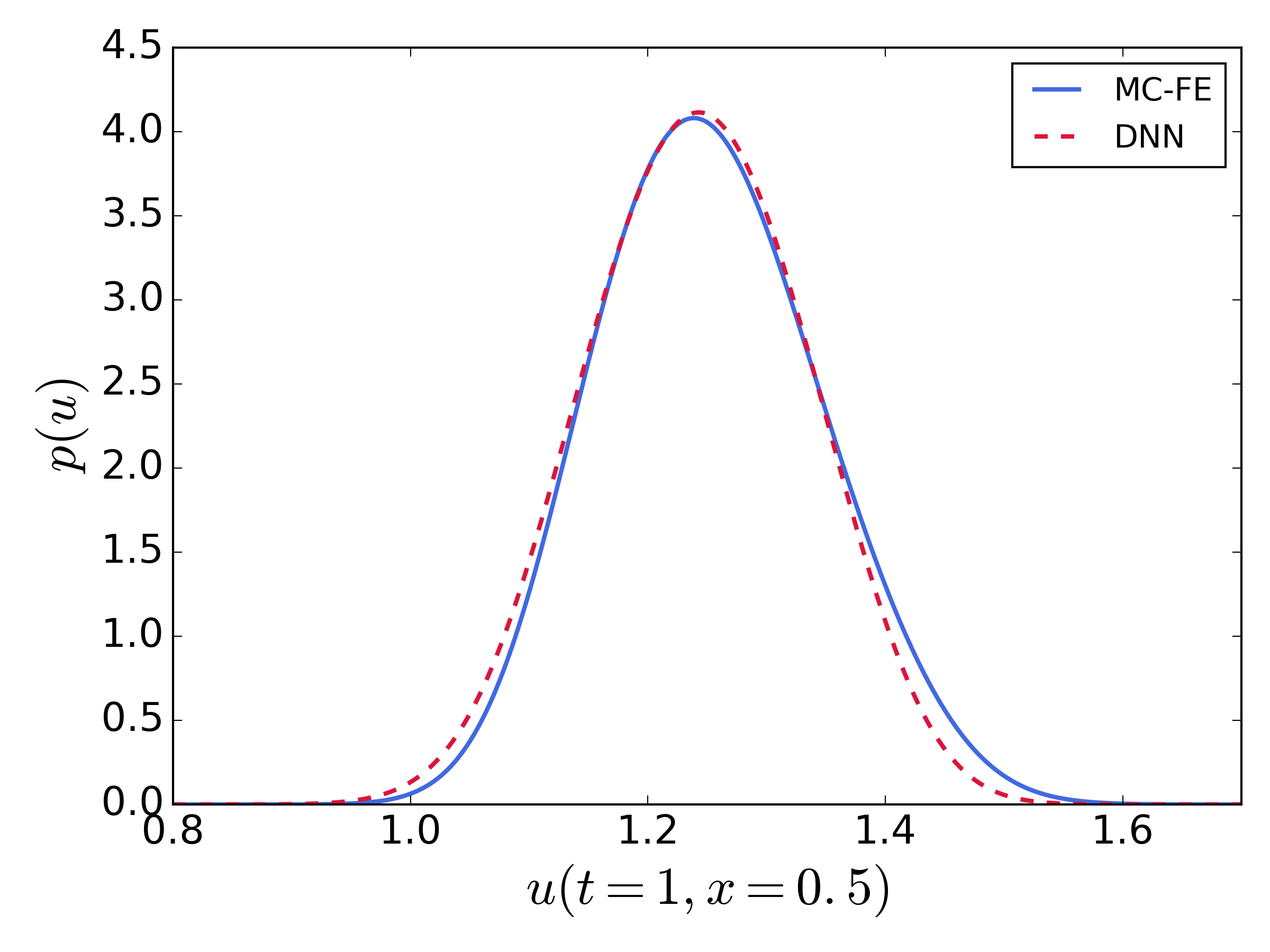}
			\caption{}
			\label{fig:diffusion2_pdf}
		\end{subfigure}
		\captionsetup{}
		\caption{A comparison between the DNN and MC-FE results for the PDF of $u$ at (t,x)=(1,0.5). The diffusion coefficient is represented by (a) Equation \ref{eqn:diffusion_coeff_1}, and (b) Equation \ref{eqn:diffusion_coeff_2}.} 
		\label{fig.diffusion_pdf}
	\end{center}
\end{figure}

\subsection{Steady heat equation with random thermal conductivity} \label{sec:example2}

\begin{figure}
	\begin{center}
		\includegraphics[width=0.75\linewidth]{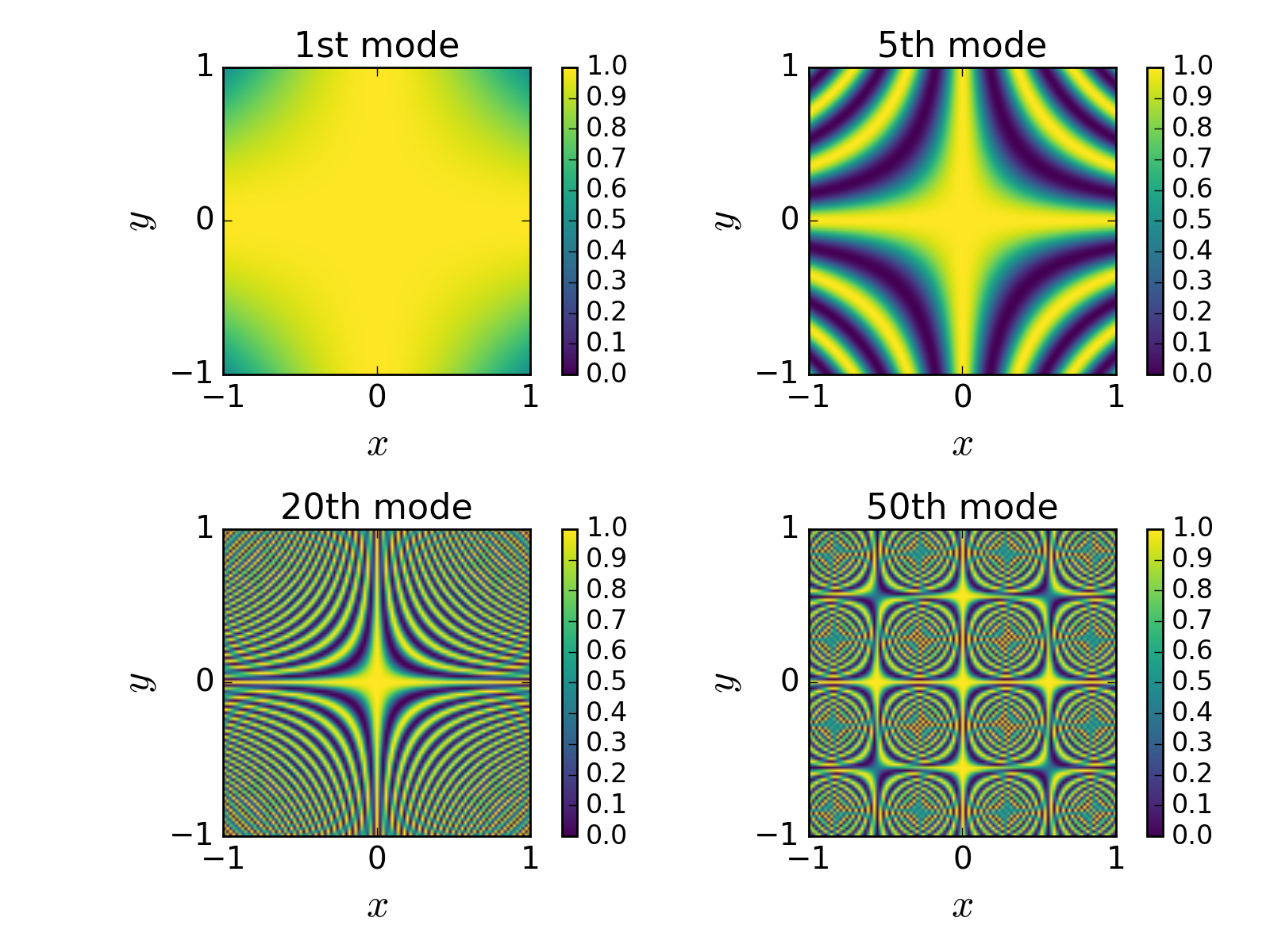}
		\caption{The first four modes of the random field represented by Equation \ref{eqn:conduction_coeff}. The decay term is not considered.} 
		\label{fig.heat_modes}
	\end{center}
\end{figure}

In this example, we consider the steady heat conduction problem on a 2-D spatial domain characterized by a random thermal conductivity coefficient, governed by the following PDE:

\begin{equation}\label{eqn:heat}
	\begin{aligned}
		-\nabla .( k( x,y,\bm{p} ) \nabla u( x,y,\bm{p} ))&=f( x,y ), \; \; \; x,y\in\left [ -1,1 \right ],\bm{p} \in \left [ 0,1 \right ]^{d}, \\
		u( -1,y,\bm{p} )= u( 1,y,\bm{p} )&= u( x,-1,\bm{p} )= u( x,1,\bm{p} )=0, 
	\end{aligned}
\end{equation}
where $f( x,y )$ is the heat generated inside the spatial domain, set to be $f( x,y )=100 \times \left | xy \right | $, and $k( x,y,\bm{p} )$ is the thermal conductivity coefficient which has the following analytical form
\begin{equation}\label{eqn:conduction_coeff}
	k( x,y,\bm{p} )=1+\sum_{j=1}^{d}\frac{1}{j}\cos^2( \frac{\pi}{4} j^{\frac{3}{2}}xy )p_j,
\end{equation}
with $d=50$. The first four modes of this random field are depicted in figure \ref{fig.heat_modes}.

In this examples, we derive a variational form of the problem, following the discussion in section \ref{sec:variational}.  It is easy to verify that solving  the variational form of the governing PDE  in \ref{eqn:heat} is equivalent to solving the following minimization problem 

\begin{equation}\label{eqn:theta-star_variational}\begin{aligned}
		&{u^*(\bm p)}=\underset{{ u }}{\operatorname{argmin}}\, \int_{\mathcal{D}}[\frac{k}{2}((\frac{\partial u}{\partial x})^2+(\frac{\partial u}{\partial y})^2)-cu  ]\diff x \, \diff y , \quad \bm p \in  \left [ 0,1 \right ]^{d}  \\
		& \text{s.t.} \, \, u(-1,y,\bm{p})=0, u(1,y,\bm{p})=0, u(x,-1,\bm{p})=0, u(x,1,\bm{p})=0,
	\end{aligned}
\end{equation}
and following a hard assignment of constraints, we force the deep neural network surrogate to take the following form
\begin{equation} \label{trial-function-example2}
	u(x,y,\bm{p}; \bm{\theta} )=(1-x^2)(1-y^2)u_\text{DNN}(x,y,\bm{p}; \bm{\theta} ).
\end{equation}

\begin{figure}
	\begin{center}
		\includegraphics[width=0.88\linewidth]{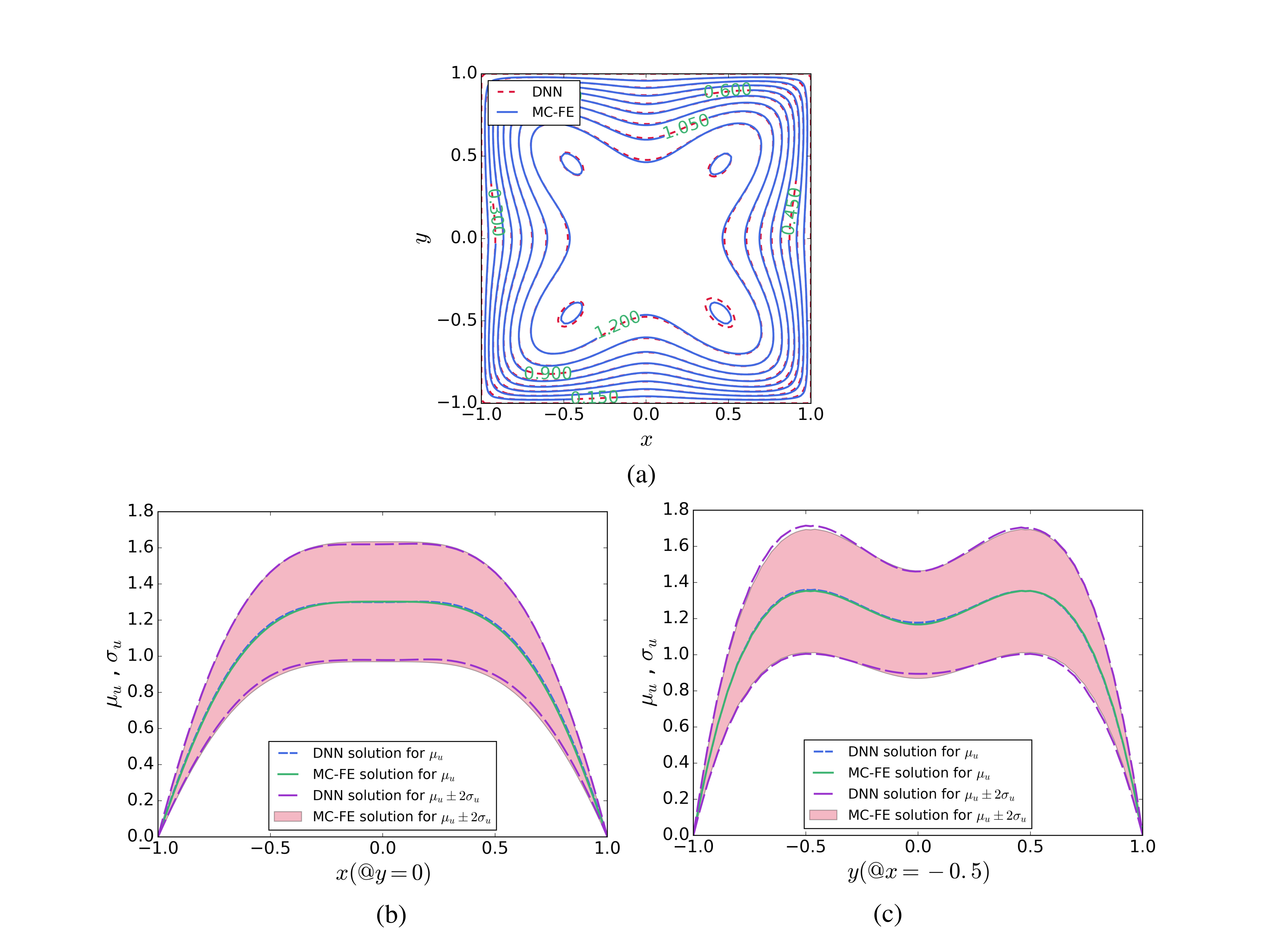}
		\caption{A comparison between the DNN and MC-FE results; (a) mean response; (b) mean and standard deviation of $u$ at $y=0$; (c) mean and standard deviation of $u$ at $x=-0.5$. } 
		\label{fig.heat_stats}
	\end{center}
\end{figure}

\begin{figure}
	\begin{center}
		\begin{subfigure}[t]{0.44 \linewidth}
			\includegraphics[width=.99\linewidth]{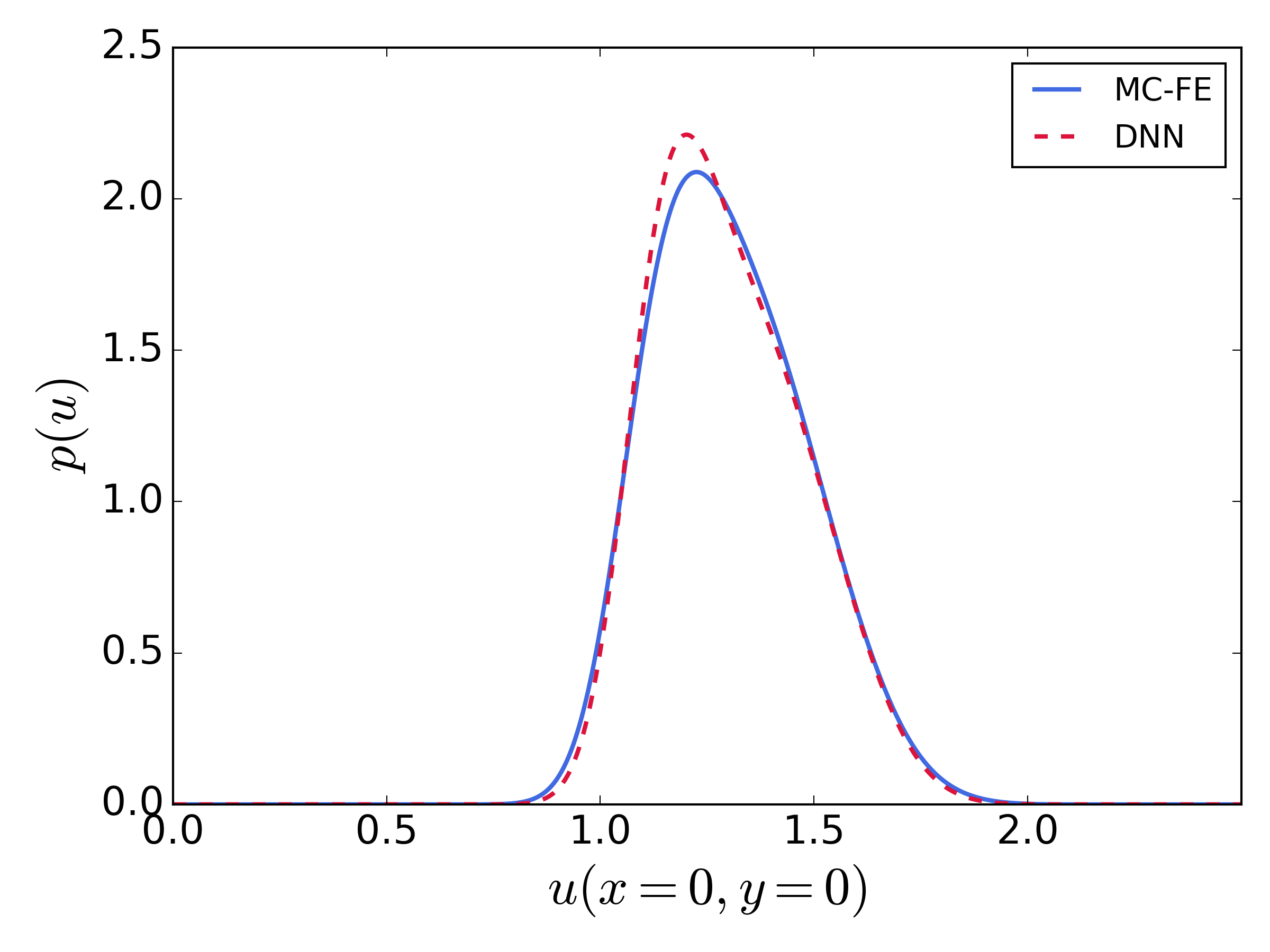}
			\caption{}
			\label{fig:heat_pdf1}	
			\centering	
		\end{subfigure}
		\quad
		\begin{subfigure}[t]{0.44 \linewidth}
			\includegraphics[width=.99\linewidth]{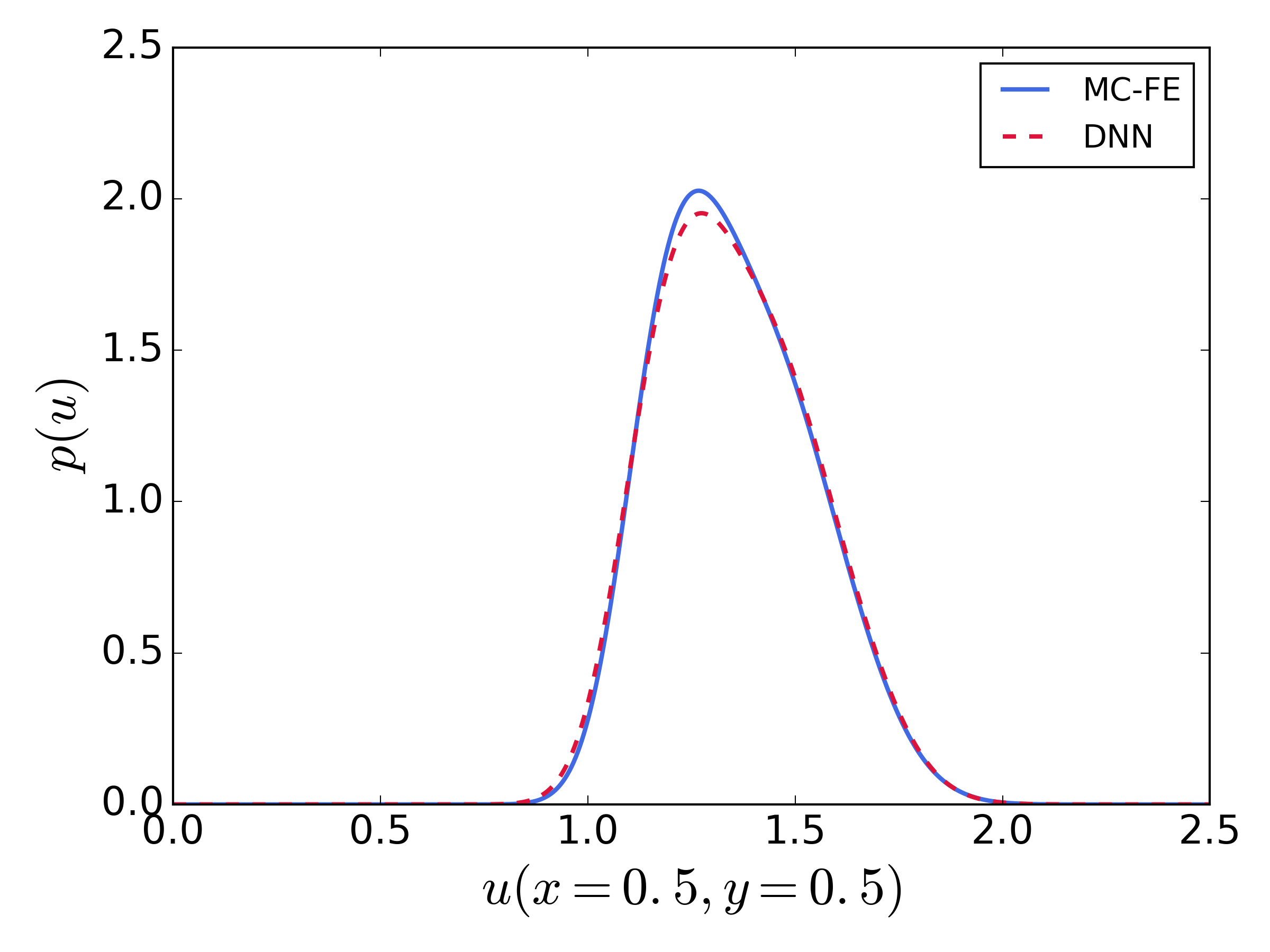}
			\caption{}
			\label{fig:heat_pdf2}
		\end{subfigure}
		\captionsetup{}
		\caption{A comparison between the deep neural network and finite element results for the PDF of $u$ at (a) $(x,y)=(0,0)$, (b) (a) $(x,y)=(0.5,0.5)$.} 
		\label{fig.heat_pdf}
	\end{center}
\end{figure}

Solution statistics over the spatial domain are represented in figure \ref{fig.heat_stats} and are compared with the MC-FE results. These figures suggest good agreement between the deep neural network and finite element results, showing the accuracy of the proposed method. Additionally, figure \ref{fig.heat_pdf} represents a comparison between the DNN and MC-FE results for probability density function of $u(x=0,y=0)$ and $u(x=0.5,y=0.5)$. For the FE solver, quadratic elements are used with a maximum mesh edge length of 0.005, which resulted in a total of 7473 nodes. Also, a total of $10^4$ MC samples are used.

\subsection{Steady heat equation with random thermal conductivity on a domain with a hole} \label{sec:example3}

\begin{figure}
	\begin{center}
		\includegraphics[width=0.95\linewidth]{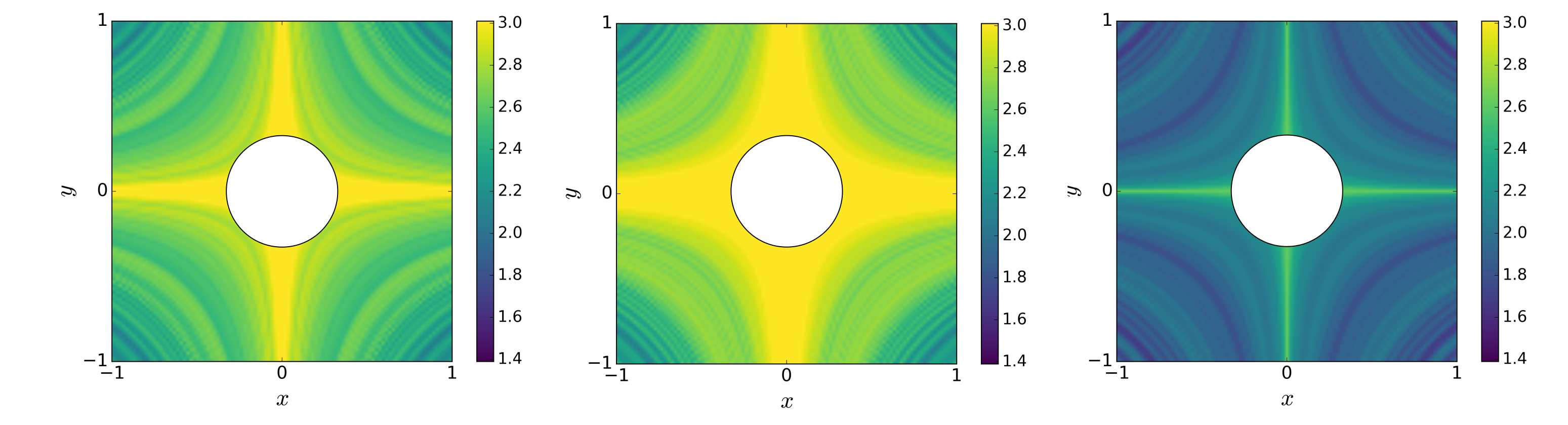}
		\caption{Geometry of the 2D plate with a hole, with three realizations of the thermal conductivity field represented by Equation \ref{eqn:conduction_coeff}. Radius of the hole is 0.3.}
		\label{fig.plate}
	\end{center}
\end{figure}

In this example, we consider the steady heat conduction problem defined by Equation \ref{eqn:heat} on a 2-D plate with a hole in the middle. The heat generated inside the spatial domain is assumed to be constant and equal to 2. $d$ is set to 30. The thermal conductivity field $k( x,y,\bm{p} )$ has the same analytical form represented by Equation \ref{eqn:conduction_coeff}, and the same zero Dirichlet boundary conditions are considered. The geometry of the 2-D plate with three realizations of the thermal conductivity field is depicted in figure \ref{fig.plate}.

Similar to the previous example, we consider a variational form of the problem, represented by Equation \ref{eqn:theta-star_variational}. Since the boundaries are rather irregular, soft assignment of constraints are considered in this example, where a penalty term is added to the loss function to account for boundary conditions, with a relatively large value for its weight, $\lambda_2=1000$.

\begin{figure}
	\begin{center}
		\includegraphics[width=0.88\linewidth]{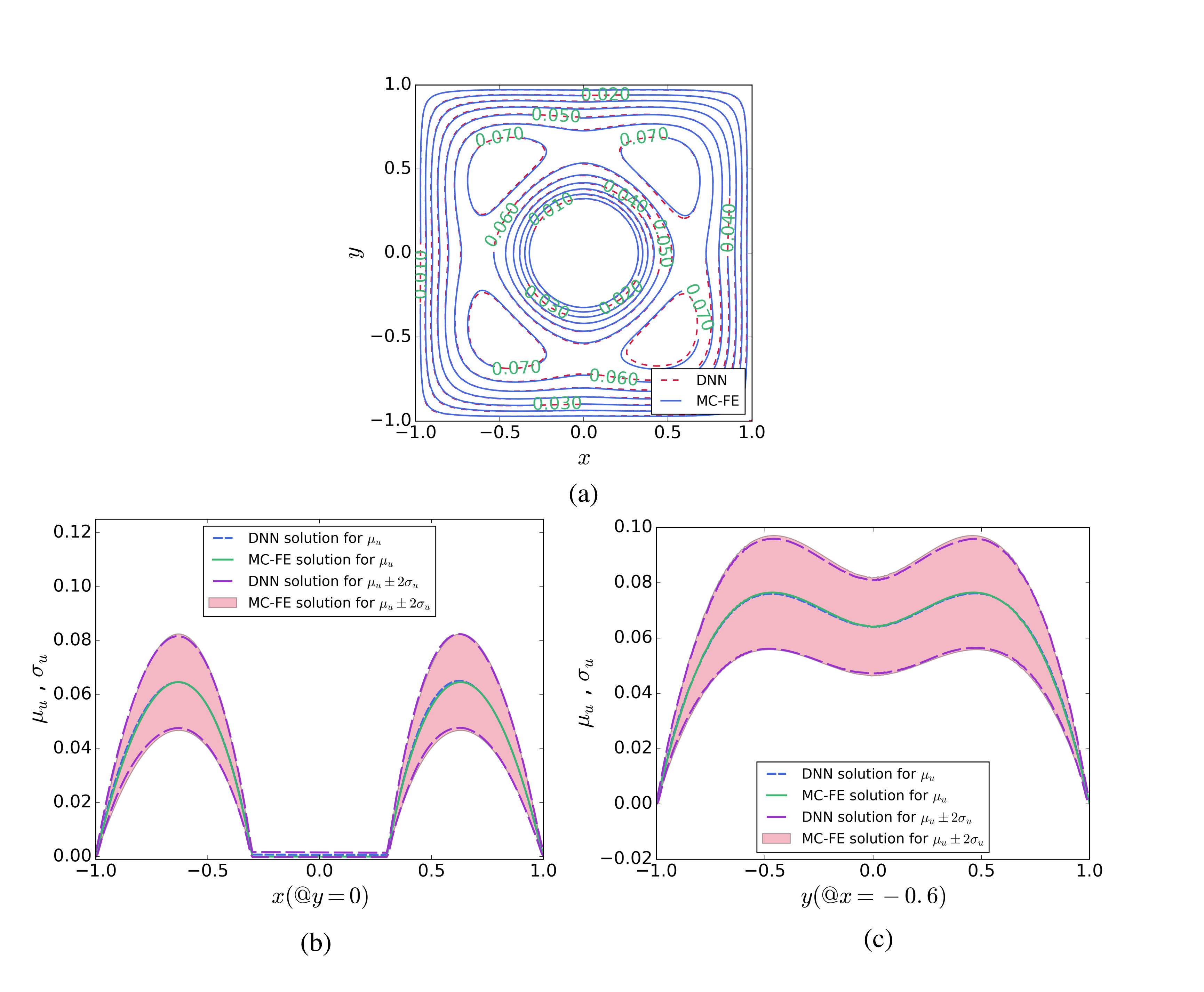}
		\caption{A comparison between the DNN and MC-FE results; (a) mean response; (b) mean and standard deviation of $u$ at $y=0$; (c) mean and standard deviation of $u$ at $x=-0.5$. The thermal conductivity field is represented by Equation \ref{eqn:conduction_coeff}.} 
		\label{fig.heat_soft_stats}
	\end{center}
\end{figure}

\begin{figure}
	\begin{center}
		\begin{subfigure}[t]{0.44 \linewidth}
			\includegraphics[width=.99\linewidth]{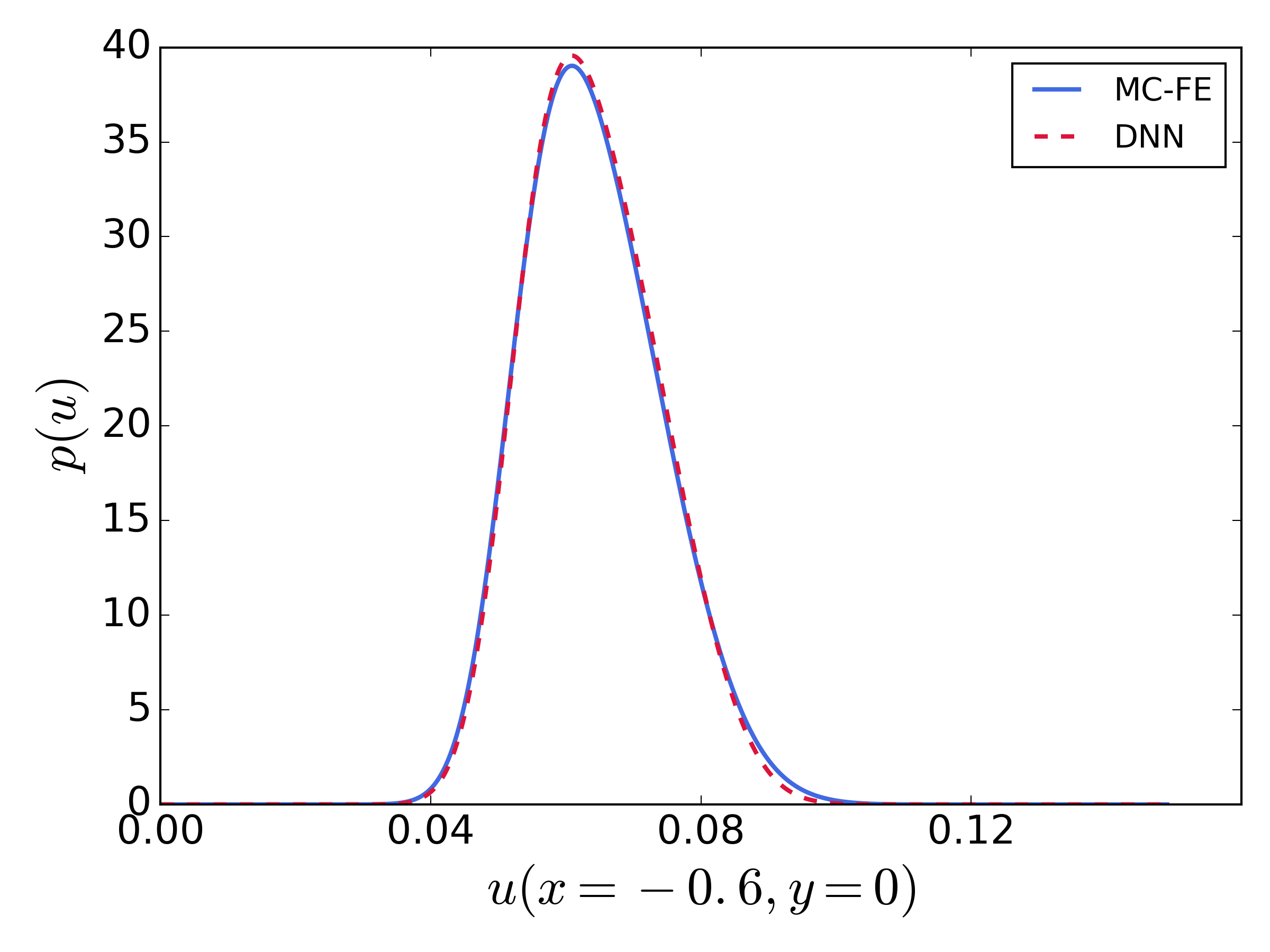}
			\caption{}
			\label{fig:heat_soft_pdf1}	
			\centering	
		\end{subfigure}
		\quad
		\begin{subfigure}[t]{0.44 \linewidth}
			\includegraphics[width=.99\linewidth]{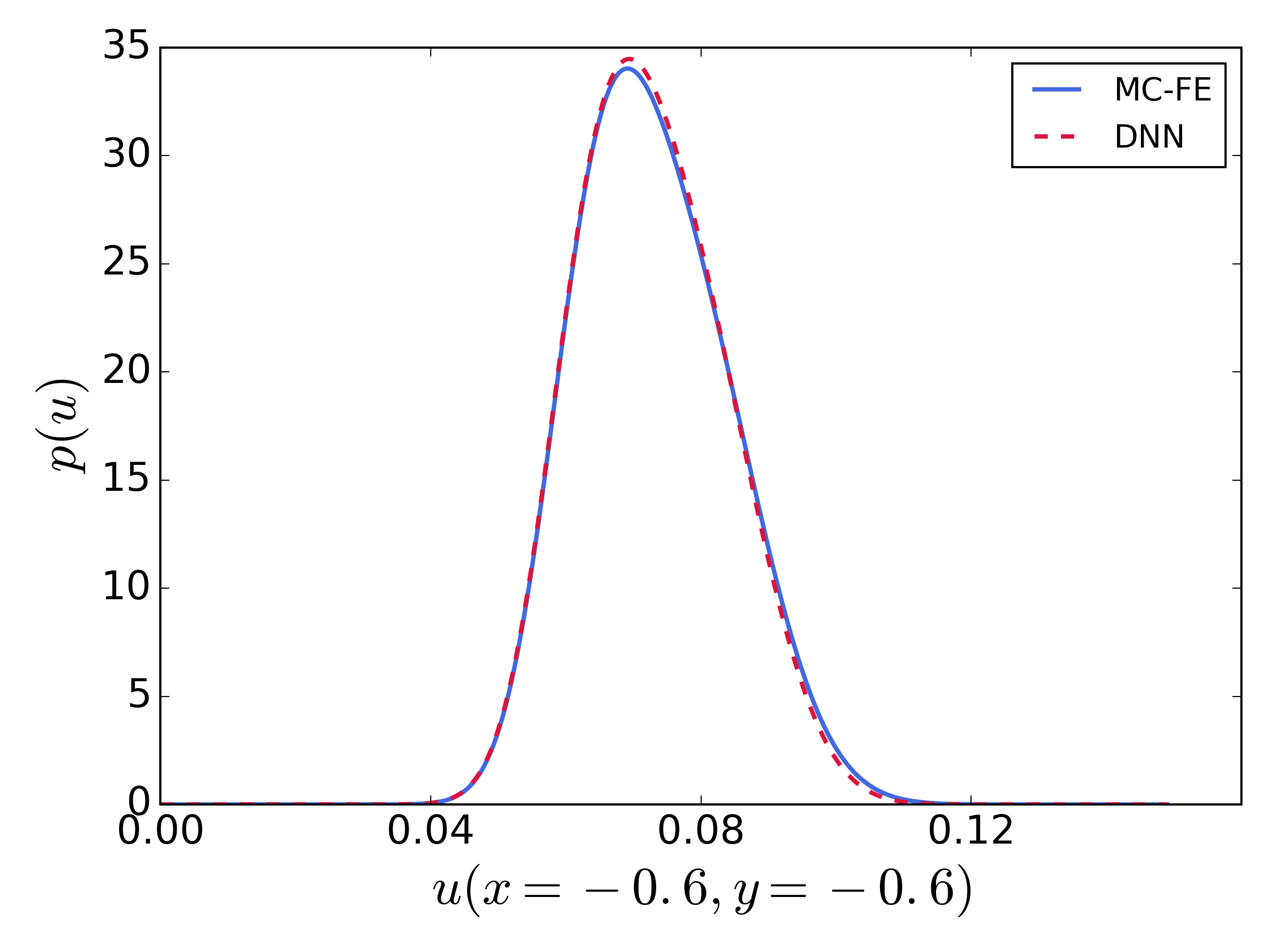}
			\caption{}
			\label{fig:heat_soft_pdf2}
		\end{subfigure}
		\captionsetup{}
		\caption{A comparison between the deep neural network and finite element results for the PDF of $u$ at (a) $(x,y)=(-0.6,0)$, (b) (a) $(x,y)=(-0.6,-0.6)$.} 
		\label{fig.heat_soft_pdf}
	\end{center}
\end{figure}

Figure \ref{fig.heat_soft_stats} shows the good agreement between the DNN and MC-FE results for the statistics of the solution. Figure \ref{fig.heat_soft_pdf} also depicts  the  probability density functions of the response at two specific locations, specifically $u(x=-0.6,y=0)$ and $u(x=-0.6,y=-0.6)$, are compared. It can be seen that the DNN is able to accurately approximate the probabilistic response.  For the MC-FE calculation, quadratic elements are used with a maximum mesh edge length of 0.002, which resulted in a total of 42972 nodes, together with a total of $10^4$ MC samples.

\section{Conclusion}\label{sec:conclusion}

We presented a new method for solving high-dimensional random PDEs using deep residual network surrogates that are trained iteratively using variants of the mini-batch gradient descent algorithm. The proposed algorithm is mesh-free over the entire spatial and temporal domains, and uses no collocation points in the stochastic space. This removes the scalability issues that collocation-based methods are experiencing with increasing number of dimensions in the stochastic space. The method can handle irregular computational domains as well. The approximate random PDE solution is computed in such a way that it will satisfy the differential operator, in strong and variational forms. In the strong form, the deep neural network parameters are trained by minimizing the squared residuals over the entire computational domain. In the variational form, a loss function is derived in such a way that its minimization is equivalent to solving the variational form of the PDE. Deep neural network parameters are then tuned by minimizing the obtained lower-order loss function. Also, the initial and boundary conditions are imposed in two different forms. In hard assignment of constraints, these conditions are inherently satisfied by enforcing the analytical form of the solution to always satisfy the initial and boundary conditions. In the soft assignment of constraints, the initial and boundary conditions are satisfied in a weak scene by adding related penalty terms to the loss function. The application of these proposed methods  is demonstrated on diffusion and heat conduction problems where the numerical results are compared with the solution obtained by finite element and Monte Carlo solutions. 

The advantages of using the proposed framework for solving random PDEs include: (1) The solution to random differential equations has a closed analytical form and is infinitely differentiable with respect to all temporal, spatial, and stochastic variables and therefore can be easily used in a variety of subsequent calculations (e.g. sensitivity analysis); (2) The proposed algorithm is embarrassingly parallel on Graphical Processing Units (GPUs); (3) The loss function is very straightforward to formulate and minimal problem-dependent setup is required before computations; (4) The method is general and can be utilized for a variety of random PDEs; (5) The solution is valid over the entire computational domain and eliminates the need for interpolation.

It should also be noted that the proposed method at its current form is not expected to dominate the best of classical numerical methods for solving random PDEs. These latter methods have been advanced over the past few decades and are optimized towards  the computational efficiency and robustness requirement in the practice. The aim of this paper is to show that DNNs can be  used to solve random PDEs accurately while offering the advantages  discussed above. The simplicity in implementation of the proposed method facilitates researchers from a wide range of scientific domains to develop, test, and analyze their ideas. In a broader context, alongside the expanding applications of machine learning methods in computational physics, we believe that our work offers an advantageous synergy between these two domains and in turn  the potential to advance both fields, which is timely considering the advances in deep learning technology, in terms of infrastructural, methodological, and algorithmic developments.

As further extensions to this work, we will pursue the following in future studies: (1) applying the method to a broader range of different types of PDEs to numerically verify that the proposed method is applicable for a wide variety of PDEs, (2) generalizing the method to be capable of handling systems of high-dimensional random PDEs; (3) investigating other efficient techniques for the enforcing initial and boundary conditions, e.g. by using Lagrange multipliers, or by developing algorithms for learning the  functions C and G  in Equation~\ref{trial-function}, and (4) investigating optimal sampling strategies in order to improve the rate of convergence.

\section*{References}

\bibliography{Nabian_Meidani}

\appendix 

\section{Derivation of the lower order loss function}
\label{app:poisson}
In what follows, on an example, we discuss how the lower order loss function can be identified.  Let us consider the Poisson equation, which is in the form of

\begin{equation} \label{Laplace}
	\frac{\partial ^2u}{\partial x^2}+\frac{\partial ^2u}{\partial y^2}+c=0,
\end{equation}
with zero Dirichlet boundary conditions.  The reduced-order variational form after using integration by parts is given by
\begin{equation} \label{Laplace_variation}
	\int_{\mathcal{D}}\left[(\frac{\partial u}{\partial x}\frac{\partial v}{\partial x}+\frac{\partial u}{\partial y}\frac{\partial v}{\partial y})-c v\right] \diff x \, \diff y =0.
\end{equation}
It can then be shown that  the functional of interest  will take the following form

\begin{equation} \label{Laplace_functional}
	F(u)=\int_{\mathcal{D}}\left[\frac{1}{2}((\frac{\partial u}{\partial x})^2+(\frac{\partial u}{\partial y})^2)-cu \right]\diff x \, \diff y 
\end{equation}
which can be easily confirmed that minimizing $F$ is equivalent to solving the Equation \ref{Laplace_variation}. 

It is difficult to formulate a general step-by-step guideline for the derivation of lower-order loss function for any given differential equations. However, in what follows, we offer few tips that can be useful  for such derivation. Let us define the functional $F$ as follows

\begin{equation} 
	F(u)=\int_{\mathcal{D}} f(x,u,\nabla u) dx,
\end{equation}
for some function $f$. In order to minimize the functional, we need that for any $v \in \mathcal{V}$, 

\begin{equation} 
	\lim_{\epsilon\to 0} \frac{F(u+\epsilon v)-F(u)}{\epsilon}=0.
\end{equation}
By applying a Taylor expansion on $F$, we have

\begin{equation} 
	f(x,u+\epsilon v, \nabla u + \epsilon \nabla v) = f(x,u,\nabla u) + \left(\frac{\partial f (x,u,\nabla u) }{\partial u}\right) \epsilon v + \left(\frac{\partial f (x,u,\nabla u) }{\partial \nabla u}\right) \epsilon \nabla v + \mathcal{O}(\epsilon^2).
\end{equation}
Therefore, we have

\begin{equation} 
	\lim_{\epsilon\to 0} \frac{F(u+\epsilon v)-F(u)}{\epsilon}=\int_{\mathcal{D}} \left( \frac{\partial f}{\partial u}v +\frac{\partial f}{\partial \nabla u} \nabla v \right) dx.
\end{equation}
By using integration by parts, we get 

\begin{equation} 
	\lim_{\epsilon\to 0} \frac{F(u+\epsilon v)-F(u)}{\epsilon}=\int_{\mathcal{D}} \left( \frac{\partial f}{\partial u} - \nabla . \frac{\partial f}{\partial \nabla u} \right) v dx.
\end{equation}
Since the left-hand side is supposed to be zero for all $v \in \mathcal{V}$, it is required that

\begin{equation} 
	\frac{\partial f(x,u,\nabla u)}{\partial u} - \nabla . \left( \frac{\partial f (x,u,\nabla u)}{\partial \nabla u} \right) = 0,
\end{equation}
which is known as the \emph{Euler-Lagrange} equation for the functional $F$.

With a known functional $F$, i.e. a known function $f$, one can use the Euler-Lagrange equation to derive a governing PDE.    Now, in order to derive the functional associated with a known PDE (which can be regarded as a Euler-Lagrange equation), we have to take the  steps  described above in the reverse direction. There are a number of rules which can simplify the process of finding the functional. For instance, assuming $f$ is independent of $x$, the following rules may be reversed and used:

\noindent 1. If $f$ consists of  $|\nabla u|^2$, the Euler-Lagrange equation will have the term   $-2 \Laplace u$.

\noindent 2. If $f$ consists of  $|u|^p, \forall p \ge 2$, the Euler-Lagrange equation will have the term $p|u|^{p-2}u$.

\noindent 3. If $f$ consists of  a term $g(u)$, where $g$ is a function of one variable, the Euler-Lagrange equation will have the term ${g}'(u)$.

\end{document}